\documentclass{article}

\PassOptionsToPackage{authoryear, round}{natbib}
\usepackage[preprint]{neurips_2026}

\usepackage[utf8]{inputenc} % allow utf-8 input
\usepackage[T1]{fontenc}    % use 8-bit T1 fonts
\usepackage{hyperref}       % hyperlinks
\usepackage{url}            % simple URL typesetting
\usepackage{booktabs}       % professional-quality tables
\usepackage{amsfonts,amsmath}       % blackboard math symbols
\usepackage{nicefrac}       % compact symbols for 1/2, etc.
\usepackage{microtype}      % microtypography
\usepackage{xcolor}         % colors

\usepackage{algorithm}
\usepackage{algpseudocode}
\usepackage{caption}
\usepackage{wrapfig}
\usepackage{graphicx}
\graphicspath{{figures/}}
\usepackage{rotating}
\usepackage{multirow}
\usepackage{array}

\usepackage{float}
\title{Discrete Diffusion for Complex and Congested Multi-Agent Path Finding with Sparse Social Attention}

\author{%
  \bfseries Yuanzhe Wang\textsuperscript{1,2,3} \quad 
  Tian Zhi\textsuperscript{1}\thanks{Corresponding author.} \quad 
  Zihang Wei\textsuperscript{1,3} \quad 
  Hongguang Wang\textsuperscript{1,3} \quad 
  Jiaming Guo\textsuperscript{1} \\[0.4em]
  \bfseries Yang Zhao\textsuperscript{4} \quad 
  Zisheng Liu\textsuperscript{1,3} \quad 
  Shiyu Quan\textsuperscript{1,2,3} \quad 
  Xing Hu\textsuperscript{1} \quad 
  Zidong Du\textsuperscript{1} \quad 
  Yunji Chen\textsuperscript{1,3} \\[1.0em]
  \textsuperscript{1} State Key Lab of Processors, Institute of Computing Technology, CAS \\
  \textsuperscript{2} School of Advanced Interdisciplinary Sciences, CAS \\
  \textsuperscript{3} University of Chinese Academy of Sciences \\
  \textsuperscript{4} Institute of Microelectronics, CAS
}

\begin{document}

\maketitle

\begin{abstract}
Multi-Agent Path Finding (MAPF) is a coordination problem that requires computing globally consistent, collision-free trajectories from individual start positions to assigned goal positions under combinatorial planning complexity. In dense environments, suboptimal initial plans induce compound conflicts that hinder feasible repair. For repair-based solvers like LNS2, initial plan quality critically affects downstream repair, yet this factor remains underexplored. We propose DiffLNS, a hybrid framework that integrates a discrete denoising diffusion probabilistic model (D3PM) with LNS2. The D3PM serves as an initializer with sparse social attention that learns a spatiotemporal prior over coordinated multi-agent action trajectories from expert demonstrations and samples multiple joint plans. Operating directly on the categorical action space, our discrete diffusion preserves the MAPF action structure and samples from a multimodal joint-plan distribution to produce diverse drafts well suited for neighborhood repair. These drafts act as warm starts for downstream repair, which completes unfinished trajectories and resolves remaining conflicts under hard MAPF constraints. Experimental results show that despite being trained only on instances with at most 96 agents, the initializer generalizes to scenarios with up to 312 agents at inference time. Across 20 complex and congested settings, DiffLNS achieves an average success rate of 95.8\%, outperforming the strongest tested baseline by 9.6 percentage points and matching or exceeding all baselines in all 20 settings. To the best of our knowledge, this is the first work to leverage discrete diffusion for warm-starting an LNS-based MAPF solver.
\end{abstract}

\section{Introduction}

Multi-Agent Path Finding (MAPF) aims to compute collision-free trajectories that guide multiple agents from their start locations to individual goals in a shared environment. As a fundamental multi-agent coordination problem, MAPF is computationally challenging and underpins a broad spectrum of real-world applications, including warehouse automation, autonomous vehicle coordination, unmanned aircraft systems, fleet management, and multi-robot systems~\citep{stern2019multi}. Consequently, scalable suboptimal planning has remained a sustained research focus.

Among existing approaches, repair-based methods such as Large Neighborhood Search 2 (LNS2) have demonstrated strong empirical performance in practical settings. These methods start from an initial set of paths and iteratively repair selected agent subsets until conflicts are resolved~\citep{li2021lns,li2022mapf}. Recent work on LNS-based MAPF has made significant progress in the repair stage, including adaptive neighborhood selection, destroy strategies, and enhanced replanning~\citep{huang2022mlguidedlns,phan2024balance,yan2024neuralneighborhood,wang2025lns2rl}. However, the initialization stage remains comparatively underexplored, with limited work on selecting promising initial solutions~\citep{huber2025learning}. This gap is consequential because the initial plan can substantially affect the success rate and convergence speed of LNS-based repair~\citep{huber2025learning,li2021lns}. In dense and congested environments, low-quality initial plans often contain severe conflicts, deadlocks, or globally inconsistent motion patterns. As a result, the back-end planner may spend a substantial computational budget correcting poor structures rather than refining an already promising solution.

A useful initializer for LNS-based MAPF should provide globally coherent joint plans that capture meaningful spatiotemporal coordination among agents, motivating a structured generative prior over coordinated trajectories. Discrete diffusion is well suited to this formulation because it can capture complex trajectory distributions through iterative denoising, operate directly in the categorical action space, and sample diverse candidate drafts for subsequent repair~\citep{carvalho2023mpd,shaoul2025mmd,liang2025projecteddiffusion}. However, directly applying diffusion to MAPF remains challenging: the model must generate centralized joint plans for many agents, maintain temporal consistency toward individual goals, and efficiently model conflict-relevant agent--agent interactions during denoising.

To overcome the limitations of suboptimal initialization in repair-based MAPF solvers and address the challenges of applying diffusion to complex and congested MAPF environments, we introduce \emph{DiffLNS}, a hybrid framework that combines a discrete denoising diffusion probabilistic model (D3PM)~\citep{austin2021structured} with LNS2. The D3PM initializer learns a spatiotemporal prior over coordinated multi-agent action trajectories from expert demonstrations. It employs a diffusion-aware sparse social attention mechanism that dynamically constructs local neighborhoods from the current denoising state, thereby focusing computation on agents that are more likely to interact or conflict. During inference, DiffLNS samples multiple drafts for the same instance, repairs them independently with LNS2, and selects the best feasible solution. This design allows D3PM to serve not only as a learned initializer but also as a source of diverse repair seeds, making DiffLNS scalable to larger agent teams and naturally parallelizable across independently generated candidates.

We summarize our main contributions and findings below:
\begin{itemize}
\item \textbf{A hybrid framework for initialization-aware MAPF.}
We propose DiffLNS, which integrates a discrete diffusion-based generator with LNS2 repair, shifting the focus from repair-only improvement to initialization quality. To the best of our knowledge, this is the first work that leverages discrete diffusion to warm-start an LNS-based MAPF solver.

\item \textbf{Diffusion-aware sparse social attention.}
We introduce a dynamic, step-dependent sparse attention mechanism that constructs local neighborhoods from the current noisy trajectory estimate. This design reduces unnecessary interaction modeling while focusing on conflict-relevant agent pairs, improving both efficiency and plan repairability.

\item \textbf{Strong empirical generalization and performance.}
Despite being trained with at most 96 agents, DiffLNS scales to teams of up to 312 agents. Across 20 complex and congested settings, it achieves a 95.8\% average success rate and improves over LNS2+RL, the strongest tested baseline by average success rate, by 9.6 percentage points.
\end{itemize}

\section{Preliminaries}
\label{sec:preliminaries}

\subsection{Multi-Agent Path Finding}
Multi-Agent Path Finding (MAPF) is commonly defined as the problem of finding collision-free paths for multiple agents from individual start locations to individual goals in a shared graph or grid environment~\citep{stern2019multi}. Let $G \in \{0,1\}^{H \times W}$ denote the map, where $G_{h,w}=0$ indicates a traversable cell and $G_{h,w}=1$ indicates an obstacle. A set of $N$ agents is specified by start locations $\{s_i\}_{i=1}^{N}$ and goal locations $\{g_i\}_{i=1}^{N}$. The planning horizon is denoted by $T$, and $\tau \in \{0,1,\dots,T\}$ indexes discrete timesteps. At each timestep $\tau$, each agent $i$ executes an action
\[
a_i^\tau \in \mathcal{A},
\qquad
\mathcal{A} = \{\texttt{stay}, \texttt{up}, \texttt{down}, \texttt{left}, \texttt{right}\}.
\]
A valid solution must avoid vertex conflicts, where different agents occupy the same cell at the same timestep, and edge conflicts, where two agents swap positions between consecutive timesteps. Solution quality is measured by sum of costs (SOC), defined as the sum of the individual path costs over all agents.

\subsection{Discrete Denoising Diffusion Probabilistic Model}
\label{sec:prelim-d3pm}

A discrete denoising diffusion probabilistic model defines a forward Markov chain that gradually corrupts a clean discrete variable $\mathbf{x}_0$ over $K$ steps~\citep{hoogeboom2021argmax,austin2021structured}. For a one-hot row vector $\mathbf{x}_k \in \{0,1\}^C$, the one-step transition is parameterized by $\mathbf{Q}_k \in \mathbb{R}^{C\times C}$:
\begin{equation}
q(\mathbf{x}_k \mid \mathbf{x}_{k-1})
=
\mathrm{Cat}\!\left(\mathbf{x}_k;\,\mathbf{p}=\mathbf{x}_{k-1}\mathbf{Q}_k\right).
\label{eq:d3pm-forward-step}
\end{equation}
For structured discrete data, this transition is applied independently to each discrete element.

D3PM commonly adopts an $\mathbf{x}_0$-parameterization, in which the model predicts the clean state $\mathbf{x}_0$ from $\mathbf{x}_k$ through $\tilde{p}_{\theta}(\mathbf{x}_0 \mid \mathbf{x}_k)$~\citep{hoogeboom2021argmax,austin2021structured}. The reverse transition is then constructed as
\begin{equation}
p_{\theta}(\mathbf{x}_{k-1}\mid \mathbf{x}_k)
\propto
\sum_{\tilde{\mathbf{x}}_0}
q(\mathbf{x}_{k-1}, \mathbf{x}_k \mid \tilde{\mathbf{x}}_0)\,
\tilde{p}_{\theta}(\tilde{\mathbf{x}}_0 \mid \mathbf{x}_k).
\label{eq:d3pm-reverse}
\end{equation}
The standard training objective combines the variational bound with an auxiliary clean-state prediction term~\citep{austin2021structured}:
\begin{equation}
\mathcal{L}_{\lambda}
=
\mathcal{L}_{\mathrm{vb}}
+
\lambda\,
\mathbb{E}_{q(\mathbf{x}_0)}
\mathbb{E}_{q(\mathbf{x}_k \mid \mathbf{x}_0)}
\left[
-\log \tilde{p}_{\theta}(\mathbf{x}_0 \mid \mathbf{x}_k)
\right].
\label{eq:d3pm-objective}
\end{equation}
The full variational decomposition used in our implementation is provided in Appendix~\ref{app:Training Objective}. At inference time, sampling starts from $\mathbf{x}_K \sim p(\mathbf{x}_K)$ and iteratively applies the learned reverse transitions for $k=K,\dots,1$.

\subsection{Large Neighborhood Search for MAPF}
MAPF-LNS2~\citep{li2022mapf} is an unbounded-suboptimal MAPF algorithm based on large neighborhood search. It first constructs an initial plan using prioritized planning (PP)~\citep{erdmann1987multiple} with SIPPS as the low-level solver; the resulting plan may contain collisions. LNS2 then repeatedly selects a subset of agents for replanning while keeping the remaining paths fixed, and accepts an update if the number of colliding pairs does not increase. The same PP+SIPPS procedure is used for neighborhood replanning, where SIPPS computes single-agent paths under dynamic obstacles and soft collision constraints~\citep{li2022mapf}. Because each iteration repairs only a local subset of agents, the difficulty of repair depends strongly on the conflict structure of the initial plan: localized and mild conflicts are easier to resolve than heavily entangled collisions. This property makes initialization quality a key factor in the effectiveness of LNS2.

\section{Methodology}
\label{sec:methodology}
DiffLNS formulates MAPF initialization as structured generative warm-starting rather than direct feasibility solving. The denoising model learns a spatiotemporal prior over coordinated joint action sequences and generates spatiotemporally structured drafts, while LNS2 enforces hard MAPF constraints by repairing residual conflicts and incomplete paths.

\begin{figure}[htbp]
    \centering
    \includegraphics[width=\linewidth]{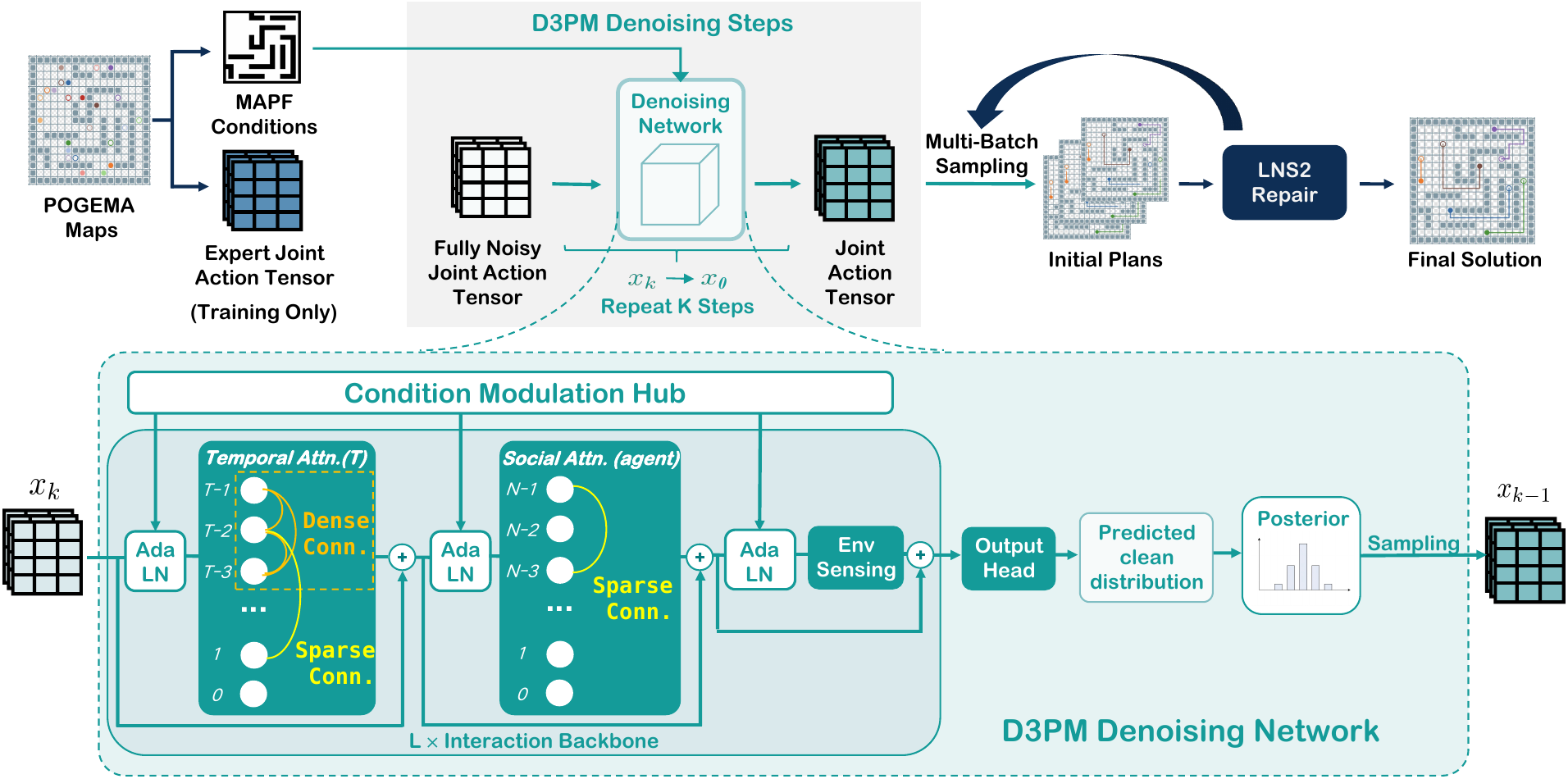}
    \caption{Overview of the DiffLNS hybrid framework. The lower panel shows the denoising network architecture used in the D3PM initializer, detailed in Section~\ref{sec:conditional-denoising-model}.}
    \label{fig:pipeline}
\end{figure}

\subsection{Hybrid Framework Overview}

\paragraph{Joint action formulation.}
An initial plan is represented as a joint discrete action tensor
\begin{equation}
\mathbf{x}_0 \in \{0,1\}^{N \times T \times C},
\label{eq:joint-action-tensor}
\end{equation}
where $N$ is the number of agents, $T$ is the planning horizon, and $C=5$ corresponds to the grid actions \{\texttt{stay}, \texttt{up}, \texttt{down}, \texttt{left}, \texttt{right}\}. Each slice $\mathbf{x}_{0,i,\tau,:}$ denotes a one-hot action token for agent $i$ at timestep $\tau$. This representation matches the categorical structure of grid-based MAPF and enables D3PM to operate directly in the discrete joint action space.

\paragraph{Overall pipeline.}
Given a MAPF instance, the D3PM initializer samples a batch of joint action drafts conditioned on the instance. Each draft is converted into a seed plan and repaired by LNS2. If one or more repaired plans are feasible, DiffLNS returns the best one according to solution quality, e.g., sum of costs (SOC). Otherwise, the system samples and repairs additional batches until the evaluation budget is exhausted. This design makes the framework complementary to repair-side improvements in LNS-based MAPF: rather than modifying the repair operator itself, DiffLNS improves the initial plans provided to it. The DiffLNS hybrid framework is illustrated in Figure~\ref{fig:pipeline}.

\subsection{Structured Initialization with Discrete Diffusion}
\paragraph{Conditional denoising model.}
\label{sec:conditional-denoising-model}
Training samples consist of MAPF conditions $\mathbf{c}$ and expert joint action tensors $\mathbf{x}_0$ defined in Eq.~\eqref{eq:joint-action-tensor}. Given a noisy sample $\mathbf{x}_k \sim q(\mathbf{x}_k \mid \mathbf{x}_0)$, we train a conditional clean-state predictor $\tilde{p}_{\theta}(\mathbf{x}_0 \mid \mathbf{x}_k,\mathbf{c})$ under the $\mathbf{x}_0$-parameterization introduced in Sec.~\ref{sec:prelim-d3pm}. The denoising network embeds the noisy action distribution of each agent at each timestep and fuses it with start, goal, map, and diffusion-step features. The resulting conditioned action tokens are processed by structured spatiotemporal blocks that integrate temporal attention, sparse social attention, and environment sensing. Temporal attention promotes motion consistency and goal progress through dense local-window attention and sparse strided global connections along each agent trajectory; sparse social attention models local conflict interactions, and environment sensing provides obstacle awareness. Details of dataset construction, preprocessing, and the full network architecture are provided in Appendix~\ref{app:training-network-details}; the sparse social attention module is described below.

\paragraph{Sparse social attention.}
\label{sec:Sparse Social Attention}
A central bottleneck in centralized denoising lies in the social interaction module. Dense all-to-all attention is not only computationally expensive but also poorly aligned with MAPF, because most conflict-relevant interactions are local, whereas attending to all agents disperses attention across many irrelevant pairs. We therefore replace dense social attention with a diffusion-aware dynamic sparse variant.

\begin{figure}[htbp]
    \centering
    \includegraphics[width=0.7\linewidth]{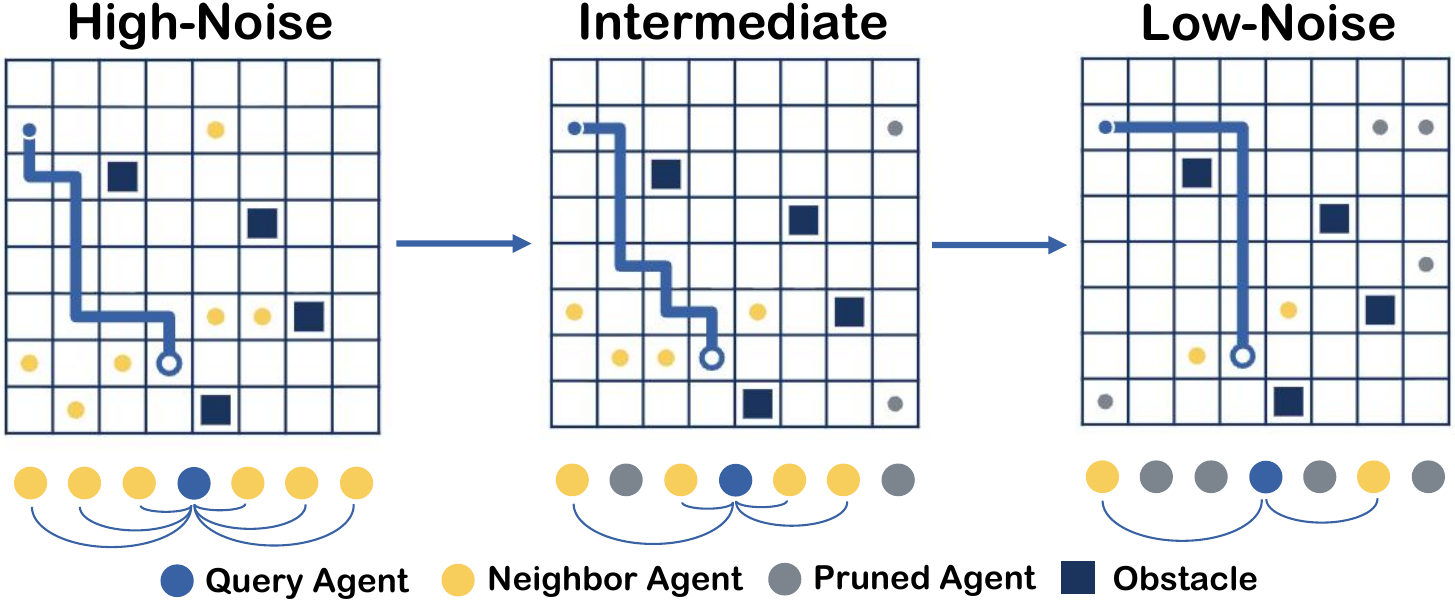}
    \caption{Diffusion-aware sparse social attention. At each denoising step, agents attend to dynamic local neighborhoods constructed from inferred-trajectory proximity, thereby focusing computation on nearby agents that are most relevant to potential conflicts.}
    \label{fig:sparse-social-attention}
\end{figure}

At diffusion step $k$, we construct the social neighborhood from the current noisy tensor $\mathbf{x}_k$ rather than from a fixed interaction graph. We first derive an inferred soft trajectory for each agent, with
$\mathbf{p}^{\mathrm{inf}}_{i,\tau}$ denoting its inferred position at timestep
$\tau$, and then compute pairwise trajectory proximity as
\begin{equation}
d_{ij}
=
\min_{\tau=1,\ldots,T}
\left\|
\mathbf{p}^{\mathrm{inf}}_{i,\tau}
-
\mathbf{p}^{\mathrm{inf}}_{j,\tau}
\right\|_1,
\label{eq:traj-proximity}
\end{equation}
which measures the minimum expected distance between two agents along the current inferred trajectories. For each agent, we retain only the top-$M_k$ nearest neighbors, where $M_k$ is set as a clipped fraction of the current team size $N$. As illustrated in Fig.~\ref{fig:sparse-social-attention}, the neighborhood is recomputed at every diffusion step: early in denoising, it remains relatively broad to preserve potentially relevant interactions, whereas later it becomes sparser as the inferred trajectory becomes more precise. Let $\mathcal{N}_k(i)$ denote the sparse neighborhood of agent $i$ at diffusion step $k$. For each timestep $\tau$, the social module first projects the agent features into query, key, and value vectors, denoted by $\mathbf{q}_{i,\tau}$, $\mathbf{k}_{j,\tau}$, and $\mathbf{v}_{j,\tau}$. Attention for agent $i$ is then computed only over $j \in \mathcal{N}_k(i)$:
\begin{equation}
\alpha_{ij}^{\tau}
=
\mathrm{softmax}_{j \in \mathcal{N}_k(i)}
\left(
\frac{\mathbf{q}_{i,\tau}^{\top}\mathbf{k}_{j,\tau}}{\sqrt{d}}
+
b_{\theta}\!\left(
\mathbf{p}^{\mathrm{inf}}_{i,\tau}
-
\mathbf{p}^{\mathrm{inf}}_{j,\tau}
\right)
\right),
\label{eq:sparse-social-attn}
\end{equation}
where $b_{\theta}(\cdot)$ is a learned geometric bias generated from relative inferred positions. The updated social feature is obtained by aggregating value vectors from the selected neighbors according to the sparse attention weights. This computation preserves focused attention on the most relevant local interactions while reducing the cost of dense $N^2$ interaction modeling.
\paragraph{Training objective.}
Our main training signal is the generative objective
\begin{equation}
\mathcal{L}_{\mathrm{gen}}
=
\mathcal{L}_{\mathrm{aux}}
+
\lambda_{\mathrm{KL}}\mathcal{L}_{\mathrm{KL}},
\label{eq:lgen}
\end{equation}
where $\mathcal{L}_{\mathrm{aux}}$ is the auxiliary clean-state prediction term in
Eq.~\eqref{eq:d3pm-objective}. The KL term corresponds to the posterior-matching
component of the variational bound $\mathcal{L}_{\mathrm{vb}}$. Let $q(\mathbf{x}_0,\mathbf{c})$ denote the expert data distribution
over clean joint action tensors and MAPF conditions. The KL term is
\begin{equation}
\mathcal{L}_{\mathrm{KL}}
=
\mathbb{E}_{q(\mathbf{x}_0,\mathbf{c})}
\mathbb{E}_{k\sim\mathrm{Unif}(\{1,\ldots,K\})}
\mathbb{E}_{q(\mathbf{x}_k\mid \mathbf{x}_0)}
\left[
\mathrm{KL}\!\left(
q(\mathbf{x}_{k-1}\mid \mathbf{x}_k,\mathbf{x}_0)
\,\|\, 
p_{\theta}(\mathbf{x}_{k-1}\mid \mathbf{x}_k,\mathbf{c})
\right)
\right].
\label{eq:kl-loss}
\end{equation}
Under the $\mathbf{x}_0$-parameterization,
$p_\theta(\mathbf{x}_{k-1}\mid \mathbf{x}_k,\mathbf{c})$ is constructed from
Eq.~\eqref{eq:d3pm-reverse} using
$\tilde{p}_{\theta}(\mathbf{x}_0 \mid \mathbf{x}_k,\mathbf{c})$. In practice,
$\mathcal{L}_{\mathrm{aux}}$ is a token-level cross-entropy loss on clean action
prediction, while $\mathcal{L}_{\mathrm{KL}}$ aligns the induced reverse transition
with the analytic posterior.

We further add a task-oriented auxiliary loss $\mathcal{L}_{\mathrm{task}}$ for goal progress, conflict reduction, and action validity, with exact forms provided in Appendix~\ref{app:Training Objective}. The final objective is
\begin{equation}
\mathcal{L}
=
\mathcal{L}_{\mathrm{gen}}
+
\mathcal{L}_{\mathrm{task}},
\label{eq:ltotal}
\end{equation}
where $\mathcal{L}_{\mathrm{gen}}$ remains the primary expert-imitation objective. For stable optimization, we first train with $\mathcal{L}_{\mathrm{gen}}$ alone and then gradually increase the weight of $\mathcal{L}_{\mathrm{task}}$.

\paragraph{Agent scalability.}
DiffLNS is not architecturally tied to a fixed team size. Although the diffusion initializer is trained on smaller teams, it can be applied to larger teams at inference time because forward corruption and reverse denoising operate tokenwise on the joint action tensor. Increasing the number of agents changes only the number of action tokens, not the form of the diffusion process.

The denoiser shares parameters across agents and is permutation-equivariant along the agent dimension:
\begin{equation}
f_{\theta}(P\mathbf{x}_k, Pc) = P f_{\theta}(\mathbf{x}_k, c),
\label{eq:perm-equiv}
\end{equation}
where $P$ is a permutation matrix acting on the agent axis. During denoising, each action token is predicted from the corrupted joint plan and the MAPF condition, using both global scene context and local interaction cues from other agents. Sparse social attention implements this locality bias through dynamic neighborhoods $\mathcal{N}_i(k)$ constructed from the current trajectory estimate. These properties make inference on larger teams well defined and provide an inductive bias for reusing local coordination patterns, but they do not guarantee distributional generalization. We therefore evaluate scaling beyond the training agent range empirically and show that DiffLNS remains effective on substantially larger teams in dense MAPF settings.

\subsection{Iterative Multi-Sample Repair with LNS2}

The diffusion initializer produces structured plans that capture useful global coordination patterns, but these plans may still contain unfinished paths, residual conflicts, or invalid moves. We therefore use LNS2 as a repair stage to convert these drafts into valid MAPF solutions.

\paragraph{Preprocessing diffusion drafts.}
Empirically, we observe that in many diffusion drafts, some agents already reach their goals, while many others terminate close to their goals. Before passing each sampled draft to LNS2, we therefore apply a short preprocessing step: redundant suffixes are removed after an agent reaches its goal, and unfinished trajectories are completed with an obstacle-aware shortest-path suffix from the last valid position to the goal. Here, invalid actions refer to action attempts that move an agent out of the map or into an obstacle. This preprocessing step preserves useful global structure while completing all trajectories to their goals before repair.

\begin{wrapfigure}{r}{0.5\linewidth}
\vspace{-1.2em}
\centering
\begin{minipage}{0.96\linewidth}
\footnotesize
\hrule
\vspace{0.25em}
\textbf{Algorithm 1} Iterative Repair with D3PM and LNS2
\vspace{0.25em}
\hrule
\vspace{0.25em}
\textbf{Input:} MAPF instance $\mathcal{I}$, initializer $p_\theta$, batch size $M$, rounds $R$, time budget $B$

\vspace{0.15em}
\begin{tabular}{@{}r@{\hspace{0.35em}}p{0.83\linewidth}@{}}
1: & \textbf{for} $r=1,\dots,R$ \textbf{do} \\
2: & \quad \textbf{if} runtime $\ge B$ \textbf{then return} failure \\
3: & \quad $\mathcal{S}_r \gets \emptyset$ \\
4: & \quad sample $\{\hat{\mathbf{x}}_0^{(m)}\}_{m=1}^{M} \sim p_\theta(\cdot\mid \mathcal{I})$ \\
5: & \quad \textbf{for} $m=1,\dots,M$ \textbf{do} \\
6: & \quad\quad \textbf{if} runtime $\ge B$ \textbf{then return} failure \\
7: & \quad\quad $\Pi^{(m)} \gets \textsc{Preprocess}(\hat{\mathbf{x}}_0^{(m)})$ \\
8: & \quad\quad $\mathbf{x}^{(m)} \gets \textsc{LNS2Repair}(\Pi^{(m)}, \mathcal{I})$ \\
9: & \quad\quad \textbf{if} feasible$(\mathbf{x}^{(m)})$ \textbf{then} $\mathcal{S}_r \gets \mathcal{S}_r \cup \{\mathbf{x}^{(m)}\}$ \\
10: & \quad \textbf{end for} \\
11: & \quad \textbf{if} $\mathcal{S}_r \neq \emptyset$ \textbf{then return} $\arg\min_{\mathbf{x}\in\mathcal{S}_r}\mathrm{SOC}(\mathbf{x})$ \\
12: & \textbf{end for} \\
13: & \textbf{return} failure
\end{tabular}
\vspace{0.25em}
\hrule
\end{minipage}
\vspace{-0.8em}
\end{wrapfigure}

\paragraph{LNS2 repair.}
Given the preprocessed draft, LNS2 performs neighborhood-based replanning to restore feasibility. Starting from the diffusion-generated trajectories, it iteratively selects subsets of agents for replanning while keeping the remaining trajectories fixed. This local repair process progressively removes residual vertex conflicts, edge-swap conflicts, and other inconsistencies until a feasible global solution is obtained or the repair budget is exhausted.

\paragraph{Multi-sample selection and iterative retry.}
For each MAPF instance, the diffusion initializer samples multiple drafts under
the same condition, allowing different global coordination patterns to be explored. If a round yields feasible repaired candidates, DiffLNS returns
the one with the smallest SOC. Otherwise, it samples a new batch and repeats the
procedure until either the maximum number of repair rounds is reached or the time budget is exhausted. This iterative strategy improves robustness by exploring diverse structured initializations within the given evaluation budget.

\section{Experiments}
\label{sec:main-experiment}
\paragraph{Baselines and metrics.}
We compare DiffLNS with LNS2~\citep{li2022mapf}, LNS2+RL~\citep{wang2025lns2rl}, HMAGAT~\citep{jain2026hmagat}, and LaCAM3~\citep{Okumura2024}. These baselines cover a strong repair-based solver, a learning-augmented repair method, a state-of-the-art learning-based solver, and a strong classical anytime solver that can continue improving SOC after finding a feasible solution within the time budget. For DiffLNS, we generate candidates in batches, with $M=4$ diffusion samples per batch, and use a 120\,s downstream LNS2 repair budget for each candidate. DiffLNS terminates successfully once any candidate is repaired into a feasible solution. DiffLNS, LNS2, LNS2+RL, and HMAGAT are evaluated as fixed-budget methods: for each benchmark family and agent cardinality, they use the same per-setting time limit, and timeout before finding a feasible solution is counted as failure. By contrast, LaCAM3 is evaluated under a matched per-instance wall-clock budget, with its time limit set to the actual runtime of DiffLNS on the same instance. Since this protocol differs from the fixed-budget setting, LaCAM3 is used only as a matched-time reference for DiffLNS and is not directly compared with the other baselines. For a fair compute protocol, learning-based methods with neural inference, including DiffLNS, LNS2+RL, and HMAGAT, are evaluated with access to the same NVIDIA L40S GPU, while all CPU-based components, including LNS2 repair and classical search baselines, are run in the same CPU environment. We report success rate (SR), average sum of costs (SOC) over successful instances, and average runtime over all instances using actual elapsed time, including failed runs. HMAGAT terminates when either the wall-clock limit or the 512-step episode limit is reached. DiffLNS runtime includes both diffusion generation and downstream LNS2 repair. Detailed evaluation protocols are provided in Appendix~\ref{app:main-result-details}.

\paragraph{Evaluation settings.}
All evaluation instances are randomly generated with POGEMA~\citep{skrynnik2025pogema}. We consider five environment families: \emph{Small Random} ($10\times 10$, obstacle density 0.175), \emph{Medium Maze} ($25\times 25$, average obstacle density 0.33), \emph{Medium Room} ($23\times 23$, average obstacle density 0.34), \emph{Medium Warehouse} ($25\times 25$, obstacle density 0.35), and \emph{Large Maze} ($33\times 33$, average obstacle density 0.33). These settings emphasize dense and congested MAPF regimes, where global coordination and repair quality become major bottlenecks. Detailed per-setting statistics and more result analyses are provided in Appendix~\ref{app:main-result-details}.

\subsection{Results}

\begin{figure*}[htbp]
\centering
\includegraphics[width=\textwidth]{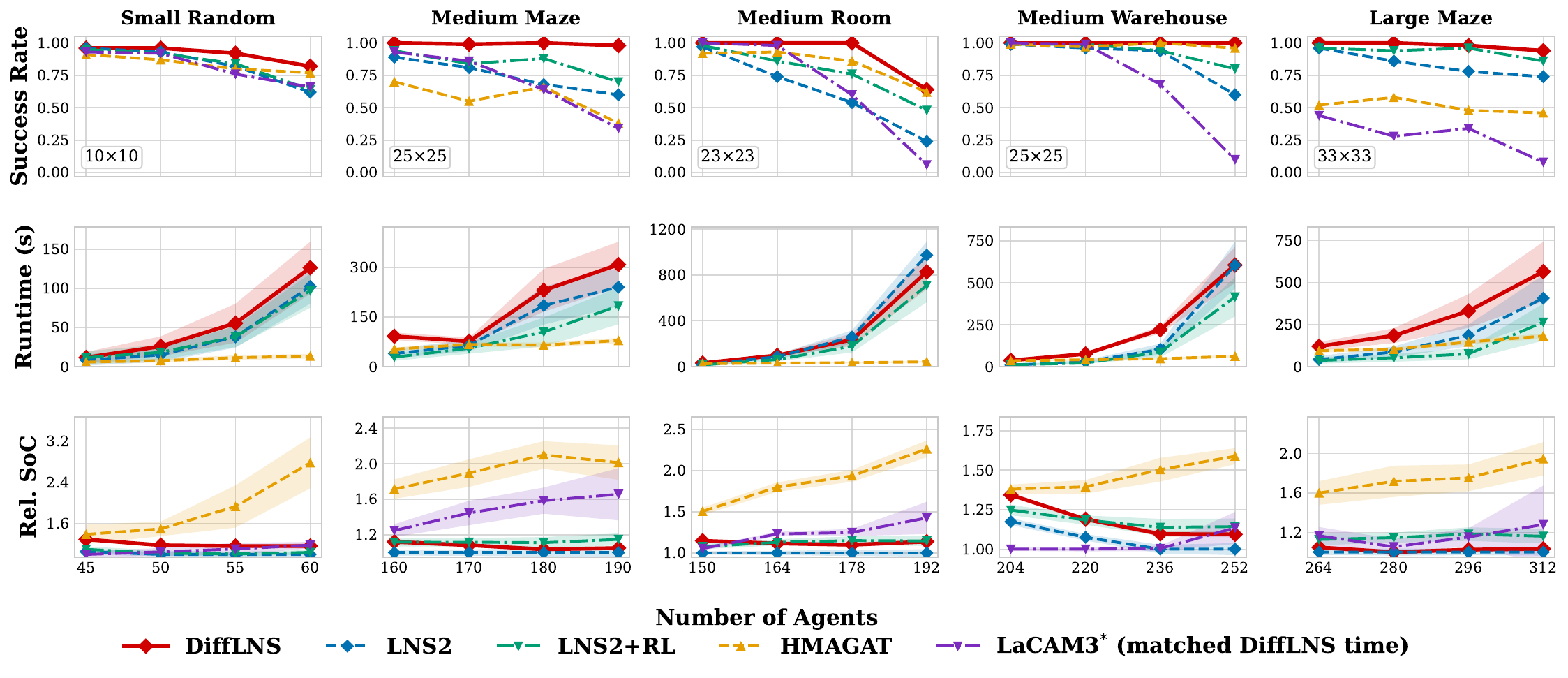}
\caption{
Performance comparison on MAPF benchmarks across different environments.
The bottom row reports relative SOC, where each value is normalized by the lowest SOC among all methods for the same environment and agent number. Transparent regions indicate 95\% confidence intervals.
}
\label{fig:mapf_results}
\vspace{-.1cm}
\end{figure*}

As shown in Fig.~\ref{fig:mapf_results}, DiffLNS matches or exceeds all baselines in SR across 20 evaluated settings, with a 95.8\% average SR and a 9.6-point gain over LNS2+RL, the strongest baseline by average SR. The advantage of DiffLNS is especially clear in maze and warehouse environments, where the success rates of classical baselines decrease more sharply as congestion increases. Under the matched-time protocol, DiffLNS achieves higher 
SR than LaCAM3 in all dense settings, especially at larger agent counts, although it has higher SOC in some scenarios. HMAGAT is fast and stable in some room and warehouse cases, but its SOC is consistently higher, indicating less cost-efficient coordination. Its runtime should also be interpreted with its 512-step episode limit: HMAGAT stops when either the wall-clock or episode limit is reached, so unfinished rollouts may terminate early and would yield extremely large SOC if evaluated directly. Overall, these results show that DiffLNS is robust across agent scales and support the effectiveness of combining generative initialization with classical repair.

DiffLNS incurs higher average runtime in most evaluated settings because each candidate requires reverse-diffusion denoising followed by LNS2 repair. Runtime tends to increase with instance difficulty, as harder cases require more generated candidates and repair attempts before a feasible solution is found; detailed candidate statistics are provided in Appendix~\ref{app:candidate-statistics}. Since candidates are independent, denoising and repair can be parallelized with sufficient GPU and CPU resources, which can substantially reduce wall-clock runtime. DiffLNS therefore trades additional but parallelizable computation for higher feasibility in congested settings while maintaining competitive SOC.

\subsection{Initialization Comparison under Fixed LNS2 Repair}
\label{sec:Initialization-Comparison}

\begin{figure}[htbp]
    \centering
    \includegraphics[width=\linewidth]{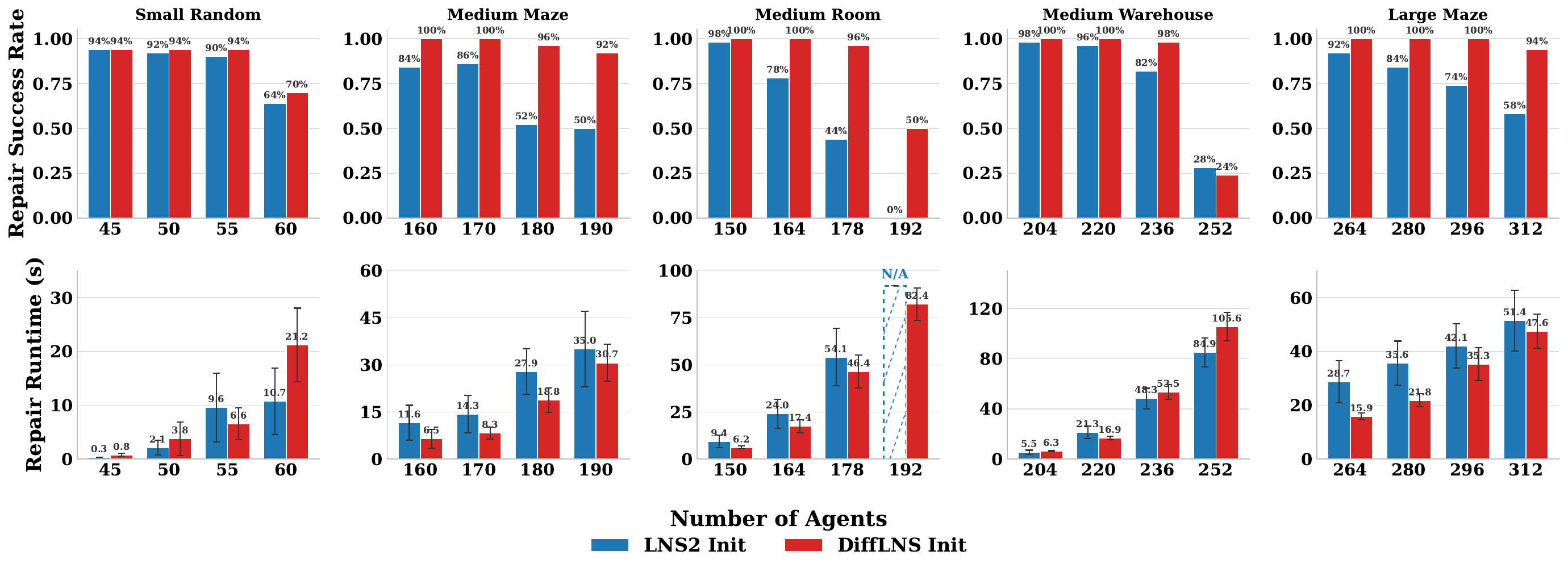}
    \caption{Initialization comparison under fixed LNS2 repair across five benchmark families. Both initialization strategies are repaired by the same LNS2 procedure under the same repair budget. Runtime is computed over successful repairs only, with N/A denoting no successful repairs. Error bars indicate 95\% confidence intervals.}
    \label{fig:repairability_compare}
\end{figure}

To isolate the effect of initialization quality from the full DiffLNS pipeline, we disable multi-sample generation and compare a single diffusion-generated initialization with the official LNS2 initialization. Both initializations are passed to the same downstream LNS2 repair procedure under the same 120\,s repair budget. Figure~\ref{fig:repairability_compare} reports the repair success rate (SR) and the LNS2 repair runtime, where runtime is computed only over successful repairs and excludes the cost of initialization generation.

Figure~\ref{fig:repairability_compare} shows that diffusion-generated initializations improve repair success in most dense and congested settings, with larger gains in highly coupled maze and room scenarios, while often leading to comparable or lower LNS2 repair time among successfully repaired instances. Overall, even a single diffusion-generated initialization provides a more repairable starting point for LNS2, indicating that DiffLNS gains are not solely due to multi-sample generation and selection. Appendix~\ref{app:pp-multistart-congested} further shows that D3PM warm-starting outperforms repeated prioritized-planning initializations under the same multi-start budget.

Interestingly, higher repairability does not necessarily require fewer raw conflicts before repair. In our experiments, DiffLNS initializations usually contain more vertex or edge conflicts than the official LNS2 initialization, yet often achieve higher repair success with comparable or lower repair time. This is consistent with prior work on selecting promising initial solutions for LNS-based MAPF~\citep{huber2025learning}: initialization quality should be judged by downstream LNS outcomes rather than by pre-repair statistics alone. The official LNS2 initialization uses sequential PP+SIPPS, which can produce locally low-conflict and low-cost paths under a fixed priority order but may yield globally rigid plans that are difficult to reorganize through local repair. In contrast, D3PM samples joint action plans from a learned expert trajectory prior, producing more coordinated and structured drafts that can be easier for LNS2 to repair even without fewer raw conflicts.

\subsection{Ablation Study on Sparse Social Attention}
To evaluate the proposed sparse social attention, we compare it with a dense all-to-all variant while keeping the rest of the DiffLNS framework unchanged. We focus on the densest settings of three representative benchmark environments: Medium Maze ($N=190$), Medium Room ($N=192$), and Medium Warehouse ($N=252$), where social interaction modeling is most critical.

\begin{figure}[htbp]
    \centering
    \includegraphics[width=0.75\linewidth]{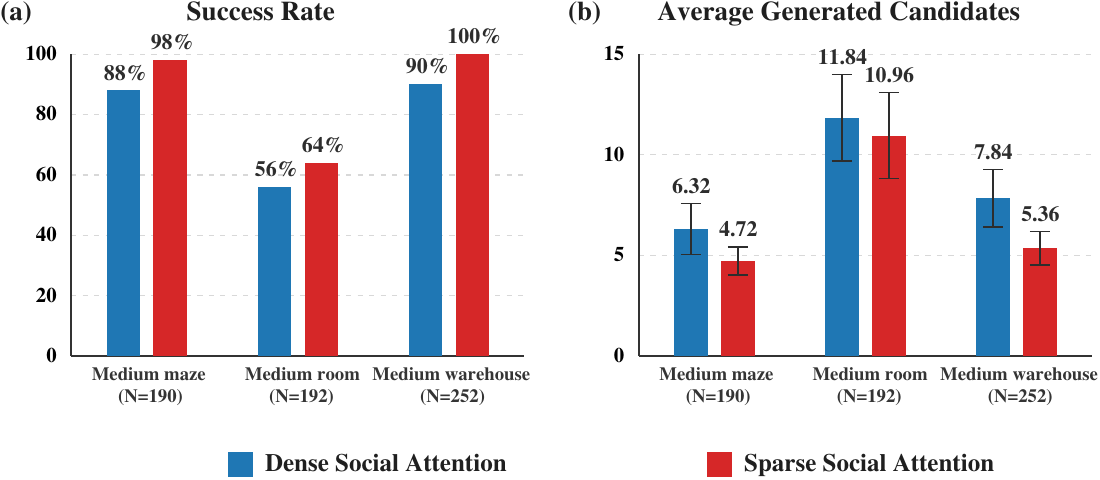}
    \caption{Ablation of sparse versus dense social attention on three dense maps. (a) Repair success rate. (b) Average number of generated candidates. Error bars indicate 95\% confidence intervals.}
    \label{fig:attention_ablation}
\end{figure}

As shown in Fig.~\ref{fig:attention_ablation}, sparse social attention consistently achieves higher repair success rates while requiring fewer generated candidates across all three settings. Averaged over the three tested settings, the dense variant also incurs approximately 12\% higher end-to-end runtime than the sparse variant. In congested MAPF instances, conflict-relevant interactions are primarily local, whereas dense attention distributes modeling capacity across many irrelevant agent pairs. By dynamically focusing on nearby agents that are most relevant to the current trajectory estimate, the sparse variant provides more effective social context and produces initial plans that are easier for LNS2 to repair in highly congested scenarios.

\subsection{Limitations}
\label{sec:limitations}

DiffLNS is most useful on dense, difficult MAPF instances where initialization quality strongly affects repair; on simpler instances, pure LNS2 is often sufficient, making the extra initialization cost less necessary. Since the current denoiser is trained on $23 \times 23$ maps, generalization to substantially larger maps may be affected by distribution shift and require broader training scales. Moreover, DiffLNS remains constrained by the downstream LNS2 repair stage: results on the hardest \emph{Medium Room} setting and the fixed-repair analysis in Section~\ref{sec:Initialization-Comparison} show that D3PM warm-starting improves repairability but cannot fully overcome extremely difficult repair cases under a limited budget.

\section{Related Work}

\paragraph{Classical MAPF solvers.}
Classical MAPF algorithms formulate planning as combinatorial search. Representative methods include conflict-based search (CBS), bounded-suboptimal variants such as EECBS, and joint-configuration search methods such as LaCAM~\citep{sharon2012cbs,li2021eecbs,okumura2023lacam}. Large neighborhood search (LNS) offers a scalable repair-based paradigm that starts from an initial plan and improves it by replanning agent subsets while keeping the remaining paths fixed~\citep{li2021lns,li2022mapf}. Although initialization quality becomes increasingly important on harder instances~\citep{li2021lns}, most LNS-based MAPF studies focus on improving the repair process, including neighborhood selection, neural neighborhood generation, adaptive destroy strategies, and reinforcement-learning-based replanners~\citep{huang2022mlguidedlns,yan2024neuralneighborhood,phan2024balance,wang2025lns2rl}. Direct studies of MAPF initialization remain limited and mainly select from existing solutions rather than generating higher-quality initializations~\citep{huber2025learning}. Our work instead improves the quality and diversity of initial plans provided to LNS2, targeting a complementary stage of the LNS pipeline.

\paragraph{Learning-based MAPF.}
Learning-based MAPF methods amortize coordination through neural policies or predictors. Early methods such as PRIMAL, DHC, and SCRIMP learn decentralized local coordination through imitation learning or multi-agent reinforcement learning~\citep{Sartoretti2019Primal,ma2021dhc,wang2023scrimp}. Recent methods improve interaction modeling with graph, hypergraph, or token-based architectures, including MAGAT, HMAGAT, MAPF-GPT, and centralized predictors such as RAILGUN~\citep{li2021magat,jain2026hmagat,andreychuk2025mapfgpt,tang2025railgun}. Although efficient at inference time, these methods typically produce limited rollouts and struggle to strictly satisfy hard MAPF constraints in dense environments. In contrast, DiffLNS generates multiple globally structured joint plans as initialization candidates and relies on LNS2 repair to guarantee feasibility.

\paragraph{Diffusion and generative planning.}
Diffusion models have been widely adopted as trajectory priors for robot planning, motion generation, and multi-robot coordination, often in continuous state spaces with additional collision handling or planner guidance~\citep{carvalho2023mpd,shaoul2025mmd,liang2025projecteddiffusion,liang2025dgd}. Their stochastic sampling naturally yields diverse candidates for a given problem instance. However, most diffusion-based planners operate over continuous trajectories, whereas grid-based MAPF is discrete and requires strict vertex and edge conflict constraints. We therefore formulate diffusion over discrete joint action tensors and use its samples to warm-start LNS2, combining multimodal generative priors with classical hard-constraint repair.

\section{Conclusion}
We presented DiffLNS, a hybrid MAPF framework that uses D3PM as a learned initializer for LNS2. DiffLNS models joint action plans through a discrete diffusion process and generates structured repair seeds conditioned on the MAPF instance, shifting the focus from repair-only improvement to initialization quality. These drafts are not required to satisfy all hard constraints by themselves; instead, they provide coordinated spatiotemporal priors that make downstream LNS2 repair more robust. Together with diffusion-aware sparse social attention and multi-sample repair, this design improves repair success in dense and congested settings while maintaining competitive solution quality. More broadly, our results suggest that discrete diffusion models can serve as structured, iterative generators for multi-agent decision making, providing useful joint priors for downstream planning and control.

{
\small

% \bibliographystyle{plainnat}
% \bibliography{references}

}

%%%%%%%%%%%%%%%%%%%%%%%%%%%%%%%%%%%%%%%%%%%%%%%%%%%%%%%%%%%%
\newpage
\appendix
\section{Outline}

This appendix provides additional results, protocol details, and implementation
details that complement the main paper. Appendix~\ref{app:pp-multistart-congested}
presents a controlled PP-Multistart comparison that separates the effect of the
learned diffusion initializer from the benefit of repeated LNS2 initialization
attempts. Appendix~\ref{app:main-results} reports the complete main experimental
results, including benchmark details, evaluation budgets, candidate-generation
statistics, and per-setting numerical results. Appendix~\ref{app:main-result-observations}
provides additional observations on the behavior of DiffLNS and the compared
baselines across scene families. Finally, Appendix~\ref{app:training-network-details}
describes the implementation details of the training pipeline and denoising
network, including expert-data construction, the conditional denoising
architecture, supervised losses, the training schedule, and optimization
hyperparameters.

\section{Additional Experiment}
\label{app:pp-multistart-congested}

We further examine whether the improvement of DiffLNS can be explained by
repeated LNS2 initialization attempts. To this end, we construct a PP-Multistart baseline under a controlled fixed
candidate budget, while replacing the diffusion initializer with the official
prioritized-planning initialization used by LNS2. We evaluate both methods on
the most congested setting of each benchmark family: \emph{Small Random} with
$60$ agents, \emph{Medium Maze} with $190$ agents, \emph{Medium Room} with
$192$ agents, \emph{Medium Warehouse} with $252$ agents, and \emph{Large Maze}
with $312$ agents. This experiment is separate from the fixed-time main evaluation and is designed to isolate the effect of initializer quality under the same number of candidate
attempts. Both methods use the same candidate-generation budget: batch
size $4$, at most $5$ batches, and at most $20$ generated candidates. We report
success rate (SR), average sum of costs (SOC), average runtime over successful
instances, and the average number of generated candidates.

\begin{figure}[htbp]
    \centering
    \includegraphics[width=\linewidth]{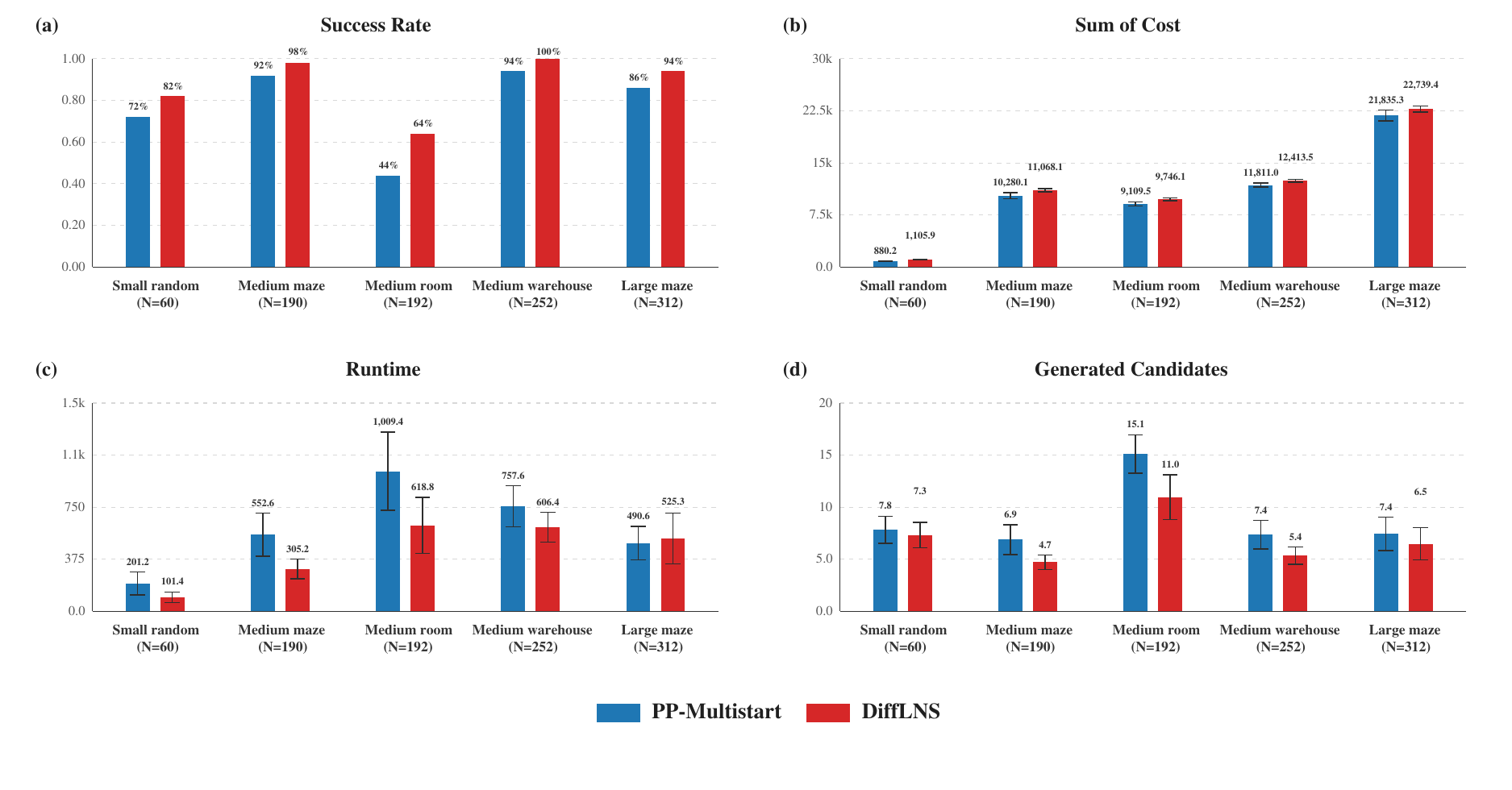}
    \caption{
    Comparison between DiffLNS and PP-Multistart on the most congested setting
    of each benchmark family. Both methods use the same multi-round
    candidate-generation budget. Error bars indicate 95\% confidence intervals
    over test instances.
    }
    \label{fig:pp-multistart-comparison}
\end{figure}

Figure~\ref{fig:pp-multistart-comparison} shows that increasing the number of
prioritized-planning initializations is not sufficient to match the robustness
of DiffLNS in highly congested scenarios. Under the same multi-start budget,
DiffLNS achieves higher SR on all five congested settings. The largest
improvement appears on \emph{Medium Room} with $192$ agents, where DiffLNS
improves SR from $44\%$ to $64\%$. DiffLNS also improves SR from $72\%$ to
$82\%$ on \emph{Small Random}, from $92\%$ to $98\%$ on \emph{Medium Maze},
from $94\%$ to $100\%$ on \emph{Medium Warehouse}, and from $86\%$ to $94\%$
on \emph{Large Maze}. These results indicate that learned diffusion
initialization provides more reliable repair seeds than repeated
prioritized-planning restarts, with the advantage becoming especially important
when PP-Multistart exhibits a clear drop in success rate.

The SOC results show a different trend from SR. PP-Multistart obtains slightly
lower SOC than DiffLNS on all five settings, suggesting that
prioritized-planning restarts can produce shorter solutions when they succeed.
However, SOC is averaged only over successful instances. Since DiffLNS solves
more instances, the successful set of DiffLNS can include harder cases that
PP-Multistart fails to solve. These additional solved cases often require longer
coordinated paths and can increase the average SOC. Therefore, the lower SOC of
PP-Multistart should not be interpreted as uniformly better solution quality,
but rather as a result computed on a smaller and easier subset of instances.
The absolute SOC values also increase on larger and more congested maps,
reflecting longer path lengths and greater coordination difficulty.

The runtime and candidate statistics further highlight the
robustness--efficiency trade-off. The reported runtime includes both
initialization and repair. PP-Multistart generally requires more generated
candidates than DiffLNS. For example, it uses $15.1$ candidates on average on
\emph{Medium Room}, compared with $11.0$ for DiffLNS, and it also uses more
candidates on \emph{Medium Maze}, \emph{Medium Warehouse}, and \emph{Large Maze}.
Despite this larger candidate usage, PP-Multistart still obtains lower SR,
suggesting that repeated prioritized-planning initializations are less reliable
repair seeds than diffusion-generated initializations. Runtime follows a similar
pattern in most settings, with PP-Multistart requiring longer total runtime on
\emph{Small Random}, \emph{Medium Maze}, \emph{Medium Room}, and
\emph{Medium Warehouse}. The exception is \emph{Large Maze}, where
PP-Multistart is slightly faster, but still has lower SR and uses more
candidates.

These results indicate that the benefit of DiffLNS is not merely due to using
more initialization attempts. Under the same multi-start budget, the diffusion
initializer provides more structured and repairable starting plans than repeated
prioritized-planning initializations, leading to higher repair robustness in
dense MAPF instances while using fewer candidates in most settings.

\section{Complete Main Experiment Results}
\label{app:main-results}

\subsection{Supplementary Experimental Details}
\label{app:main-result-details}

This appendix supplements the main experimental section with additional protocol
details and per-setting statistics for interpreting
Table~\ref{tab:appendix-main-results}, Fig.~\ref{fig:mapf_results}, and
Fig.~\ref{fig:repairability_compare}. Table~\ref{tab:appendix-map-details}
summarizes the benchmark families, map sizes, obstacle-density ranges, and
evaluated agent cardinalities. Table~\ref{tab:appendix-budget-details} reports
the number of evaluated instances and the fixed evaluation budgets for the
fixed-budget baselines.

\paragraph{Baselines and implementation details.}
The main experiments compare DiffLNS with four MAPF baselines: 
LNS2~\citep{li2022mapf}\footnote{\url{https://github.com/Jiaoyang-Li/MAPF-LNS2}}, 
LNS2+RL~\citep{wang2025lns2rl}\footnote{\url{https://github.com/marmotlab/LNS2-RL}},
HMAGAT~\citep{jain2026hmagat}\footnote{\url{https://github.com/proroklab/hmagat}}, 
and LaCAM3~\citep{Okumura2024}\footnote{\url{https://github.com/Kei18/lacam3}}. 
LNS2 is a strong classical repair-based solver. LNS2+RL is
a learning-augmented variant of LNS2. HMAGAT is a state-of-the-art
learning-based MAPF solver; in our evaluation, we use its k-means variant and
set the maximum episode length to 512 steps. LaCAM3 is a strong classical
anytime solver. After finding an initial feasible solution, LaCAM3 continues to
improve the solution cost until the given time limit is reached.

For DiffLNS, we generate candidates in batches, with $M=4$ diffusion samples per
batch. The downstream LNS2 repair budget for each DiffLNS candidate is fixed to
120\,s. DiffLNS terminates successfully once any candidate is repaired into a
feasible solution. For DiffLNS, LNS2, LNS2+RL, and HMAGAT, we use fixed
per-setting time limits that increase with scene difficulty, ranging from
180\,s to 1200\,s, as reported in Table~\ref{tab:appendix-budget-details}. An
instance is counted as a failure for all fixed-budget methods if the time limit
is reached before a feasible solution is found. For all LNS-based methods,
including DiffLNS, LNS2, and LNS2+RL, the repair neighborhood size is fixed to 8.

\paragraph{Licenses of existing assets.}
We use publicly available implementations only for evaluation and baseline comparison. MAPF-LNS2 is released under the USC Research License; LNS2+RL is released under the MIT License; HMAGAT builds on MAGAT, whose included implementation is released under the MIT License; LaCAM3 is released under the MIT License; and POGEMA is released under the MIT License. We cite the corresponding papers and provide the official repository links above.

\paragraph{Hardware and parallelization.}
All methods were evaluated on machines equipped with NVIDIA L40S GPUs and
Intel Xeon Gold 6348 CPUs at 2.60\,GHz. Evaluation instances were parallelized
across independent worker processes, with each process handling one MAPF
instance at a time. Each worker was allocated 8 logical CPU threads. Methods
requiring GPU acceleration were assigned one NVIDIA L40S GPU per worker, while
CPU-only methods used the same 8-thread CPU setting without GPU acceleration.
This protocol ensures that all methods are evaluated under the same
per-instance CPU resource allocation, with GPU resources provided only to
methods that require neural inference.

\paragraph{Evaluation metrics and runtime protocol.}
We report success rate (SR), average sum of costs (SOC), and average runtime. SR is the percentage of instances in which all agents reach their assigned goals within the given budget. SOC is computed only over successful instances, while runtime is averaged over all instances using the actual elapsed time of each run, including failed runs. For fixed-budget solvers, failed runs may terminate either at the wall-clock time limit or at a method-specific stopping condition. In particular, HMAGAT terminates when either the wall-clock limit or the 512-step episode limit is reached, whichever comes first. For DiffLNS, runtime includes both diffusion generation and downstream LNS2 repair.

DiffLNS, LNS2, LNS2+RL, and HMAGAT are evaluated under fixed per-setting time
budgets, where each setting is defined by a benchmark family and an agent
cardinality. These budgets are assigned according to setting difficulty estimated
from preliminary pilot runs: larger or more congested settings are given longer
budgets, while easier settings use shorter budgets to avoid unnecessary
computation. Within each setting, all fixed-budget methods use the same time
limit, and timeout before finding a feasible solution is counted as failure.

LaCAM3 is evaluated under a matched per-instance time budget rather than this
fixed-budget protocol. Specifically, for each instance, the time limit of
LaCAM3 is set to the actual runtime of DiffLNS on the same instance. This
protocol allows LaCAM3 to be used as a matched-time reference for DiffLNS in SR
and SOC. Since this protocol differs from the fixed-budget setting, LaCAM3 is
not directly compared with the other fixed-budget methods, and its runtime is
omitted from the runtime comparison.

\begin{figure}[htbp]
    \centering
    \includegraphics[width=0.8\linewidth]{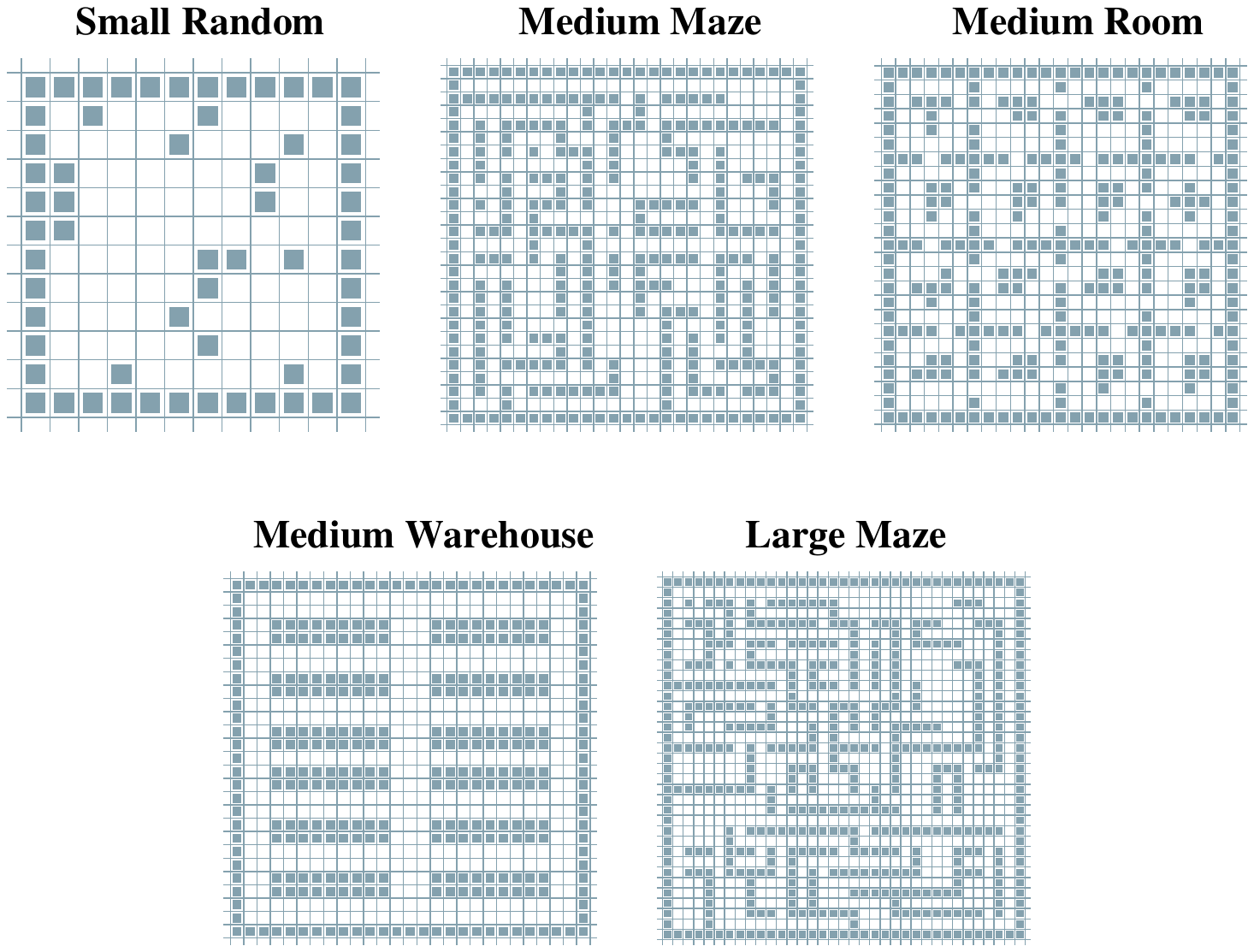}
    \caption{Example test maps generated with POGEMA for the evaluated benchmark
    families. These maps illustrate the obstacle layouts and structural
    differences across the random, maze, room, and warehouse environments used
    in our experiments.}
    \label{fig:test-scene-maps}
\end{figure}

\paragraph{Benchmark statistics.}
All evaluation instances are randomly generated with 
POGEMA~\citep{skrynnik2025pogema}\footnote{\url{https://github.com/CognitiveAISystems/pogema}}.
Example test maps from the evaluated benchmark families are shown in
Fig.~\ref{fig:test-scene-maps}. The reported agent densities are computed with
respect to free cells, i.e., as the ratio between the number of agents and the
number of non-obstacle cells. The instance counts shown in
Table~\ref{tab:appendix-budget-details} are the actual numbers of evaluated MAPF
instances for each agent cardinality. The listed time limits correspond to the
fixed-budget methods, namely DiffLNS, \textsc{LNS2}, \textsc{LNS2+RL}, and
\textsc{HMAGAT}. LaCAM3 is excluded from this table because it is evaluated
under matched per-instance budgets determined by the runtime of DiffLNS on the
same instance.

\begin{table}[htbp]
\centering
\small
\setlength{\tabcolsep}{7pt}
\renewcommand{\arraystretch}{1.12}
\caption{Benchmark families used in the main experiments. Obstacle-density
ranges are measured from the generated benchmark instances.}
\label{tab:appendix-map-details}
\begin{tabular}{lccc}
\toprule
\textbf{Map} & \textbf{Size (H $\times$ W)} & \textbf{Obstacle Density} & \textbf{Agent Cardinalities} \\
\midrule
Small Random     & $10\times10$ & $17.5\%$ & 45 / 50 / 55 / 60 \\
Medium Maze      & $25\times25$ & $27.4$--$36.5\%$ & 160 / 170 / 180 / 190 \\
Medium Room      & $23\times23$ & $31.9$--$35.0\%$ & 150 / 164 / 178 / 192 \\
Medium Warehouse & $25\times25$ & $34.6\%$          & 204 / 220 / 236 / 252 \\
Large Maze       & $33\times33$ & $29.3$--$36.8\%$ & 264 / 280 / 296 / 312 \\
\bottomrule
\end{tabular}
\end{table}

\begin{table}[t]
\centering
\small
\setlength{\tabcolsep}{7pt}
\renewcommand{\arraystretch}{1.12}
\caption{Evaluation budgets for the main experiments. Levels 1--4 correspond to
the four agent cardinalities listed in Table~\ref{tab:appendix-map-details},
from low to high density. Each entry reports the number of cases and the
corresponding time limit. The listed time limits apply to the fixed-budget
methods (DiffLNS, \textsc{LNS2}, \textsc{LNS2+RL}, and \textsc{HMAGAT}).}
\label{tab:appendix-budget-details}
\begin{tabular}{lcccc}
\toprule
\textbf{Map} & \textbf{Level 1} & \textbf{Level 2} & \textbf{Level 3} & \textbf{Level 4} \\
\midrule
Small Random &
100 cases / 180\,s &
100 cases / 180\,s &
100 cases / 180\,s &
100 cases / 240\,s \\

Medium Maze &
100 cases / 240\,s &
100 cases / 240\,s &
50 cases / 480\,s &
50 cases / 480\,s \\

Medium Room &
100 cases / 240\,s &
100 cases / 240\,s &
50 cases / 480\,s &
50 cases / 1200\,s \\

Medium Warehouse &
100 cases / 240\,s &
100 cases / 240\,s &
50 cases / 480\,s &
50 cases / 1200\,s \\

Large Maze &
50 cases / 360\,s &
50 cases / 360\,s &
50 cases / 600\,s &
50 cases / 1200\,s \\
\bottomrule
\end{tabular}
\end{table}

\subsection{Candidate Generation Statistics}
\label{app:candidate-statistics}

Table~\ref{tab:generation-by-scene-density} reports the average number of
generated candidates used by DiffLNS under the iterative inference strategy.
As agent density and coordination difficulty increase, DiffLNS typically needs
more candidate generations to obtain a feasible repaired solution, which
accounts for part of the runtime increase observed in the main results.

\begin{table}[t]
\centering
\caption{Average number of generated candidates used by DiffLNS. For each
scenario family, Levels 1--4 denote increasing agent density.}
\label{tab:generation-by-scene-density}
\begin{tabular}{lcccc}
\toprule
Scenario & Level 1 & Level 2 & Level 3 & Level 4 \\
\midrule
Small random & 4.64 & 4.68 & 5.52 & 7.32 \\
Medium maze & 4.00 & 4.16 & 4.24 & 4.72 \\
Medium room & 4.00 & 4.00 & 4.16 & 10.96 \\
Medium warehouse & 4.00 & 4.00 & 4.00 & 5.36 \\
Large maze & 4.00 & 4.08 & 4.96 & 6.48 \\
\bottomrule
\end{tabular}
\end{table}

% The full numerical table is unchanged except for the caption.
\begin{sidewaystable*}[p]
\centering
\scriptsize
\setlength{\tabcolsep}{4pt}
\renewcommand{\arraystretch}{1.08}
\caption{Complete main experiment results. Higher success rate is better, while
lower SOC and runtime are better. The symbols $\uparrow$ and $\downarrow$
indicate that larger and smaller values are better, respectively. Agent density
is computed as the ratio between the number of agents and the number of free
cells, i.e., non-obstacle cells. For DiffLNS, LNS2, LNS2+RL, and HMAGAT, runtime is averaged over all evaluated instances using the actual elapsed time of each run, including failed runs. HMAGAT terminates when either the wall-clock limit or the 512-step episode limit is reached. The evaluation time limits are shown in parentheses in the runtime headers for each scene and agent cardinality. For LaCAM3, runtime is omitted because LaCAM3 is evaluated under per-instance budgets matched to DiffLNS.}
\label{tab:appendix-main-results}
\resizebox{0.96\textheight}{!}{%
\begin{tabular}{|>{\centering\arraybackslash}m{2.2cm}|*{4}{>{\centering\arraybackslash}m{1.62cm}}|*{4}{>{\centering\arraybackslash}m{1.62cm}}|*{4}{>{\centering\arraybackslash}m{1.62cm}}|}
\hline
\multirow{3}{*}{\textbf{Methods}} & \multicolumn{12}{c|}{\textbf{Small Random: $10 \times 10$ world size with 17.5\% static obstacle rate and 54.3\%(45 agents), 60.4\%(50 agents), 66.0\%(55 agents), 72.3\%(60 agents) agents density}} \\
\cline{2-13}
 & \multicolumn{4}{c|}{\textbf{Success Rate}$\uparrow$} & \multicolumn{4}{c|}{\textbf{Sum of Cost}$\downarrow$} & \multicolumn{4}{c|}{\textbf{Runtime}$\downarrow$} \\
\cline{2-13}
 & N=45 & N=50 & N=55 & N=60 & N=45 & N=50 & N=55 & N=60 & N=45 (180s) & N=50 (180s) & N=55 (180s) & N=60 (240s) \\
\hline
DiffLNS & \textbf{96\%} & \textbf{96\%} & \textbf{92\%} & \textbf{81\%} & 863.1 & 993.1 & 1,064.5 & 1,105.9 & 11.8 & 26.1 & 55.5 & 126.3 \\
LNS2    & \textbf{96\%} & 93\% & 82\% & 62\% & 705.1 & \textbf{842.3} & \textbf{915.6} & \textbf{950.3} & 8.5 & 15.1 & 38.1 & 102.5 \\
LNS2+RL & 95\% & 92\% & 84\% & 64\% & 734.6 & 863.8 & 924.5 & 988.8 & 11.6 & 18.5 & 38.3 & 97.0 \\
HMAGAT  & 91\% & 87\% & 80\% & 77\% & 928.0 & 1,260.1 & 1,766.9 & 2,639.1 & \textbf{5.9} & \textbf{7.8} & \textbf{11.6} & \textbf{13.5} \\
LaCAM3  & 93\% & 92\% & 76\% & 66\% & \textbf{668.2} & 884.9 & 1,013.0 & 1,122.9 & -- & -- & -- & -- \\
\hline

\multicolumn{1}{|c|}{} & \multicolumn{12}{c|}{\textbf{Medium Maze: $25 \times 25$ world size with 32.8\% static obstacle rate and 38.0\%(160 agents), 40.4\%(170 agents), 43.0\%(180 agents), 45.3\%(190 agents) agents density}} \\
\cline{2-13}
 & \multicolumn{4}{c|}{\textbf{Success Rate}$\uparrow$} & \multicolumn{4}{c|}{\textbf{Sum of Cost}$\downarrow$} & \multicolumn{4}{c|}{\textbf{Runtime}$\downarrow$} \\
\cline{2-13}
 & N=160 & N=170 & N=180 & N=190 & N=160 & N=170 & N=180 & N=190 & N=160 (240s) & N=170 (240s) & N=180 (480s) & N=190 (480s) \\
\hline
DiffLNS & \textbf{100\%} & \textbf{99\%} & \textbf{100\%} & \textbf{98\%} & 9,421.2 & 9,933.7 & 10,248.2 & 11,068.1 & 92.1 & 76.8 & 231.0 & 308.7 \\
LNS2    & 89\% & 81\% & 68\% & 60\% & \textbf{8,399.3} & \textbf{9,172.7} & \textbf{9,883.8} & \textbf{10,531.3} & 39.8 & 64.1 & 184.8 & 240.4 \\
LNS2+RL & 94\% & 84\% & 88\% & 70\% & 9,386.9 & 10,243.0 & 10,991.6 & 12,098.4 & \textbf{29.1} & \textbf{55.7} & 104.8 & 183.8 \\
HMAGAT  & 70\% & 55\% & 66\% & 38\% & 14,406.6 & 17,382.7 & 20,753.1 & 21,203.4 & 52.9 & 67.4 & \textbf{65.3} & \textbf{79.1} \\
LaCAM3  & 93\% & 86\% & 64\% & 34\% & 10,480.1 & 13,264.5 & 15,665.7 & 17,448.2 & -- & -- & -- & -- \\
\hline

\multicolumn{1}{|c|}{} & \multicolumn{12}{c|}{\textbf{Medium Room: $23 \times 23$ world size with 33.5\% static obstacle rate and 42.7\%(150 agents), 46.8\%(164 agents), 50.5\%(178 agents), 54.4\%(192 agents) agents density}} \\
\cline{2-13}
 & \multicolumn{4}{c|}{\textbf{Success Rate}$\uparrow$} & \multicolumn{4}{c|}{\textbf{Sum of Cost}$\downarrow$} & \multicolumn{4}{c|}{\textbf{Runtime}$\downarrow$} \\
\cline{2-13}
 & N=150 & N=164 & N=178 & N=192 & N=150 & N=164 & N=178 & N=192 & N=150 (240s) & N=164 (240s) & N=178 (480s) & N=192 (1200s) \\
\hline
DiffLNS & \textbf{100\%} & \textbf{100\%} & \textbf{100\%} & \textbf{64\%} & 7,953.4 & 8,322.1 & 8,895.6 & 9,746.1 & 31.9 & 99.5 & 235.5 & 828.0 \\
LNS2    & 97\% & 74\% & 54\% & 24\% & \textbf{6,932.2} & \textbf{7,463.4} & \textbf{8,097.5} & \textbf{8,573.7} & 16.2 & 88.2 & 259.1 & 973.2 \\
LNS2+RL & 98\% & 86\% & 76\% & 48\% & 7,464.4 & 8,424.2 & 9,308.9 & 9,822.8 & \textbf{15.7} & 62.5 & 179.2 & 709.3 \\
HMAGAT  & 92\% & 93\% & 86\% & 62\% & 10,425.7 & 13,421.1 & 15,665.3 & 19,384.4 & 26.6 & \textbf{31.0} & \textbf{35.4} & \textbf{43.0} \\
LaCAM3  & \textbf{100\%} & 98\% & 60\% & 6\% & 7,301.6 & 9,177.3 & 10,101.7 & 12,213.7 & -- & -- & -- & -- \\
\hline

\multicolumn{1}{|c|}{} & \multicolumn{12}{c|}{\textbf{Medium Warehouse: $25 \times 25$ world size with 34.6\% static obstacle rate and 49.9\%(204 agents), 53.8\%(220 agents), 57.7\%(236 agents), 61.6\%(252 agents) agents density}} \\
\cline{2-13}
 & \multicolumn{4}{c|}{\textbf{Success Rate}$\uparrow$} & \multicolumn{4}{c|}{\textbf{Sum of Cost}$\downarrow$} & \multicolumn{4}{c|}{\textbf{Runtime}$\downarrow$} \\
\cline{2-13}
 & N=204 & N=220 & N=236 & N=252 & N=204 & N=220 & N=236 & N=252 & N=204 (240s) & N=220 (240s) & N=236 (480s) & N=252 (1200s) \\
\hline
DiffLNS & \textbf{100\%} & \textbf{100\%} & \textbf{100\%} & \textbf{100\%} & 10,432.2 & 11,088.3 & 11,485.3 & 12,413.5 & 37.7 & 76.3 & 220.4 & 606.4 \\
LNS2    & 99\% & 96\% & 94\% & 60\% & 9,118.6 & 10,022.0 & \textbf{10,481.3} & \textbf{11,351.2} & \textbf{9.5} & 29.9 & 103.7 & 605.5 \\
LNS2+RL & 99\% & 99\% & 94\% & 80\% & 9,688.6 & 11,038.4 & 11,930.3 & 12,971.2 & 11.7 & \textbf{21.9} & 87.6 & 416.5 \\
HMAGAT  & 99\% & 97\% & \textbf{100\%} & 96\% & 10,716.3 & 13,005.4 & 15,755.6 & 18,028.0 & 35.9 & 41.8 & \textbf{47.9} & \textbf{62.4} \\
LaCAM3  & \textbf{100\%} & 99\% & 68\% & 10\% & \textbf{7,766.5} & \textbf{9,325.2} & 10,527.9 & 12,874.4 & -- & -- & -- & -- \\
\hline

\multicolumn{1}{|c|}{} & \multicolumn{12}{c|}{\textbf{Large Maze: $33 \times 33$ world size with 33.0\% static obstacle rate and 36.4\%(264 agents), 38.3\%(280 agents), 40.6\%(296 agents), 42.7\%(312 agents) agents density}} \\
\cline{2-13}
 & \multicolumn{4}{c|}{\textbf{Success Rate}$\uparrow$} & \multicolumn{4}{c|}{\textbf{Sum of Cost}$\downarrow$} & \multicolumn{4}{c|}{\textbf{Runtime}$\downarrow$} \\
\cline{2-13}
 & N=264 & N=280 & N=296 & N=312 & N=264 & N=280 & N=296 & N=312 & N=264 (360s) & N=280 (360s) & N=296 (600s) & N=312 (1200s) \\
\hline
DiffLNS & \textbf{100\%} & \textbf{100\%} & \textbf{98\%} & \textbf{92\%} & 19,002.0 & \textbf{20,236.1} & 21,783.6 & 22,739.4 & 121.7 & 184.8 & 331.5 & 565.8 \\
LNS2    & 96\% & 86\% & 78\% & 74\% & \textbf{18,139.4} & 20,250.2 & \textbf{21,258.9} & \textbf{22,006.9} & 43.2 & 87.9 & 186.6 & 407.6 \\
LNS2+RL & 96\% & 94\% & 96\% & 86\% & 20,457.5 & 23,201.8 & 25,140.4 & 25,551.9 & \textbf{37.3} & \textbf{53.0} & \textbf{76.5} & 265.0 \\
HMAGAT  & 52\% & 58\% & 48\% & 46\% & 29,010.3 & 34,755.6 & 37,254.9 & 42,815.3 & 95.4 & 104.0 & 146.5 & \textbf{182.3} \\
LaCAM3  & 44\% & 28\% & 34\% & 8\% & 21,133.0 & 21,323.1 & 24,504.1 & 28,135.5 & -- & -- & -- & -- \\
\hline
\end{tabular}%
}
\end{sidewaystable*}

\subsection{Additional Observations on Main Results}
\label{app:main-result-observations}

Table~\ref{tab:appendix-main-results} provides the complete scene-by-scene
numerical results underlying Fig.~\ref{fig:mapf_results}. Beyond the main
observations in the paper, several additional patterns are worth noting.

\paragraph{DiffLNS.}
The most prominent strength of DiffLNS is its consistently high success rate
across all evaluated settings. In most scene families and agent cardinality levels, it matches or exceeds all compared methods in success rate. The advantage is especially pronounced in
\emph{Medium Maze}, \emph{Medium Warehouse}, and \emph{Large Maze}, where
DiffLNS maintains nearly perfect success rates even under very dense and highly
congested conditions. This result suggests that the learned initializer is
particularly effective in scenarios where global coordination and repair
quality are the dominant bottlenecks.

In terms of solution quality, DiffLNS usually obtains slightly higher SOC than
LNS2, which is expected because LNS2 is a strong classical solver optimized for
cost efficiency on successful instances. Nevertheless, DiffLNS still achieves
competitive SOC in most settings and does not substantially sacrifice solution
quality in exchange for improved feasibility. Taken together, these results
indicate a favorable balance between robustness and cost.

Regarding runtime, DiffLNS is generally slower than the compared baselines
because each instance requires both diffusion-based generation and downstream
LNS2 repair. This overhead is discussed in the main experimental section and in
the limitations section, and it represents the main computational trade-off of
the proposed framework.

\paragraph{LNS2.}
LNS2 remains a strong classical baseline, especially in solution quality and runtime. Across many settings, it attains the lowest or near-lowest SOC and is often among the fastest fixed-budget methods. However, its success rate drops substantially as environments become more crowded. This trade-off is particularly visible in \emph{Medium Maze}, \emph{Medium Room}, \emph{Medium Warehouse}, and \emph{Large Maze}, where LNS2 can remain efficient on solved instances but fails to retain the same level of robustness as DiffLNS.

\paragraph{LNS2+RL.}
\textsc{LNS2+RL} generally improves the success rate of \textsc{LNS2} in harder
settings, indicating that learning-guided repair provides a meaningful
robustness gain. This is especially evident in the denser maze and warehouse
settings, where \textsc{LNS2+RL} narrows part of the feasibility gap to DiffLNS.
In our experiments, we adopt the best-performing hyperparameter setting reported
in the original \textsc{LNS2+RL} paper, namely neighborhood size $8$, switch
threshold $\rho=0.3$, stop-MARL-planning threshold $t_l=1.2$, and maximum
length $t_h=2.2$. However, the gain in success rate usually comes at the cost of
higher SOC than \textsc{LNS2}, and \textsc{LNS2+RL} still does not match the
strongest success-rate performance of DiffLNS in the most challenging settings.
In this sense, \textsc{LNS2+RL} can be viewed as a more robust but less
cost-efficient variant of \textsc{LNS2}. More importantly, since
\textsc{LNS2+RL} improves the repair stage whereas DiffLNS improves the
initialization stage, the two directions are largely complementary rather than
mutually exclusive. This suggests that combining a stronger learned initializer
with a stronger learned repair policy is a promising direction for further
improving success rates in highly congested MAPF settings.

\paragraph{HMAGAT.}
\textsc{HMAGAT} shows a different profile from the LNS-based methods. It is relatively stable and fast in the room and warehouse families, suggesting that its learned policy captures useful regularity in structured indoor layouts. However, \textsc{HMAGAT} also exhibits the highest SOC in nearly all scene families and struggles more severely in maze-like environments, especially \emph{Large Maze}. Its runtime should also be interpreted with its maximum episode length of 512 steps: in hard instances, unfinished rollouts can terminate early after reaching this limit, which partly explains the lower average runtime. Thus, the strength of \textsc{HMAGAT} lies more in maintaining a usable success rate in some structured layouts than in producing low-cost solutions.

\paragraph{LaCAM3.}
LaCAM3 should be interpreted separately from the other baselines because it is
evaluated under matched per-instance budgets rather than the fixed-budget
protocol. Under this matched-time setting, it remains competitive on several
easier or medium-difficulty settings and can still achieve favorable SOC on the
solved subset. However, its success rate degrades much more rapidly in dense
and difficult regimes, especially in \emph{Large Maze} and the hardest
\emph{Medium Warehouse} settings. These results suggest that although LaCAM3
remains a strong classical anytime solver, the matched-time comparison still
favors DiffLNS in highly congested instances where feasible planning under a
limited per-instance runtime becomes difficult.

\section{Training and Denoising Network Details}
\label{app:training-network-details}

This appendix provides implementation details that are omitted from the main text, including expert-data construction, the MAPF-conditioned denoising network, the exact supervised losses, the training schedule, and optimization hyperparameters. The main text defines the high-level DiffLNS framework and the generative objective; here we give the concrete choices used in our experiments.

\subsection{Training Data and Distributed Setup}

Training data are constructed from procedurally generated POGEMA~\citep{skrynnik2025pogema} MAPF instances. We use three scene families: maze, room, and warehouse. All training maps have size $23\times23$, and the dataset contains three agent cardinalities, $N\in\{32,64,96\}$. The dataset is evenly split across the three scene types and contains 4.5K instances with 32 agents, 9K instances with 64 agents, and 9K instances with 96 agents. For each instance, start and goal locations are randomly generated, and obstacle densities are sampled to produce diverse MAPF configurations. We solve these instances with LaCAM3~\citep{Okumura2024} and retain only successful rollouts as expert demonstrations.

To preserve a fixed team size within each mini-batch while training a shared denoiser across different agent counts, we use a round-robin dataset mixing strategy: each mini-batch is sampled from one fixed-$N$ subset, and the three subsets are alternated throughout each epoch. The per-GPU batch size is 8.

Training is performed with distributed data parallelism on a single node with 8 NVIDIA A100 GPUs. The total training procedure takes approximately 26 hours for 400 epochs. Random seeds are fixed to 42 for both PyTorch and NumPy. Mixed-precision training is enabled with BF16 autocasting, and activation checkpointing is used to reduce peak memory usage during backbone training.

\subsection{Model and Diffusion Configuration}
\label{app:Denoising Network Arch}

The denoising network implements the conditional clean-state predictor
$\tilde{p}_{\theta}(\mathbf{x}_0\mid \mathbf{x}_k,\mathbf{c})$ used by the
$x_0$-parameterized D3PM reverse process. It follows the general design paradigm
of diffusion transformers by using a timestep-conditioned transformer-style
backbone for denoising, but is specialized for MAPF rather than being a direct
DiT architecture~\citep{Peebles_2023_ICCV}. In particular, it operates on
structured agent-time action tokens and factorizes denoising into temporal
attention, diffusion-aware sparse social attention, and trajectory-conditioned
environment sensing.

Given a noisy joint action tensor
$\mathbf{x}_k\in[0,1]^{N\times T\times C}$, where $C=5$ corresponds to
\texttt{stay}, \texttt{up}, \texttt{down}, \texttt{left}, and \texttt{right},
the network outputs action logits
$f_{\theta}(\mathbf{x}_k,\mathbf{c},k)\in\mathbb{R}^{N\times T\times C}$.
The predicted clean distribution is
\begin{equation}
\tilde{p}_{\theta}(\mathbf{x}_0\mid \mathbf{x}_k,\mathbf{c})
=
\prod_{i=1}^{N}\prod_{\tau=1}^{T}
\mathrm{Cat}\!\left(
\mathbf{x}_{0,i,\tau,:};
\mathrm{softmax}\!\left(
f_{\theta}(\mathbf{x}_k,\mathbf{c},k)_{i,\tau,:}
\right)
\right).
\end{equation}
We use $K=100$ diffusion steps with a cosine noise schedule~\citep{pmlr-v139-nichol21a}. During training,
the diffusion step is sampled uniformly as $k\sim\mathcal{U}\{1,\ldots,K\}$.

\paragraph{Uniform forward transition matrix.}
We use the standard uniform transition kernel from the D3PM family~\citep{austin2021structured}:
\begin{equation}
\mathbf{Q}_k
=
\alpha_k \mathbf{I}
+
(1-\alpha_k)\frac{1}{C}\mathbf{1}\mathbf{1}^{\top}.
\label{eq:uniform-transition}
\end{equation}
Here $\alpha_k$ preserves the current action with probability $\alpha_k$ and redistributes the
remaining mass uniformly over the $C$ action classes. This choice is natural for MAPF because
the action space is discrete and unordered, with no meaningful notion of distance between action
classes. As the diffusion step increases, each action is therefore gradually mixed toward a uniform
categorical distribution.

\paragraph{Condition encoding.}
The MAPF condition $\mathbf{c}$ contains the obstacle map, goal map, start and
goal locations, map size, and a global density feature. Let the grid map have
width $W$ and height $H$, and denote the map-size vector by $\mathbf{m}=(W,H)$.
Let
$V_{\mathrm{free}}=\{(u,v):G_{u,v}=0\}$
denote the set of traversable cells, where $G_{u,v}=0$ indicates a free cell and
$G_{u,v}=1$ indicates an obstacle. Start and goal locations are represented in
normalized map coordinates, $\mathbf{s}_i,\mathbf{g}_i\in[-1,1]^2$. The global
agent-density feature is
\begin{equation}
\rho
=
\log\!\left(1+\frac{N}{|V_{\mathrm{free}}|}\right).
\end{equation}
The diffusion step is encoded by a sinusoidal embedding $\mathrm{PE}(k)$~\citep{NIPS2017_3f5ee243}. The
global condition is computed as
\begin{equation}
\mathbf{c}_{\mathrm{g}}
=
\psi_{\mathrm{g}}
\left(
\left[
\phi_k(\mathrm{PE}(k)),
\;
\phi_m(\log \mathbf{m})+\phi_{\rho}(\rho)
\right]
\right),
\end{equation}
where $\phi_k$, $\phi_m$, $\phi_{\rho}$, and $\psi_{\mathrm{g}}$ are learned
projections. For each agent, the agent-wise condition is
\begin{equation}
\mathbf{c}_i
=
\psi_{\mathrm{n}}
\left(
\left[
\mathbf{c}_{\mathrm{g}},
\phi_s(\mathbf{s}_i),
\phi_g(\mathbf{g}_i),
\phi_r(\mathbf{g}_i-\mathbf{s}_i)
\right]
\right).
\end{equation}
The global condition also predicts a scale gate for the map pyramid, later used
to weight multi-scale map features in Eq.~\eqref{eq:env-sensing}:
\begin{equation}
\boldsymbol{\eta}
=
\mathrm{softmax}\!\left(\psi_{\mathrm{scale}}(\mathbf{c}_{\mathrm{g}})\right),
\qquad
\boldsymbol{\eta}\in\mathbb{R}^{S}.
\label{eq:scale-gate}
\end{equation}

\paragraph{Initial denoising tokens.}
The network first embeds each noisy action distribution and adds start, goal, and relative-goal
embeddings. The initial denoising token for agent $i$ at timestep $\tau$ is
\begin{equation}
\mathbf{x}^{(0)}_{i,\tau}
=
\phi_a(\mathbf{x}_{k,i,\tau,:})
+
\phi_s(\mathbf{s}_i)
+
\phi_g(\mathbf{g}_i)
+
\phi_r(\mathbf{g}_i-\mathbf{s}_i),
\qquad
\mathbf{x}^{(0)}_{i,\tau}\in\mathbb{R}^{D}.
\label{eq:initial-denoising-token}
\end{equation}
Here the superscript in $\mathbf{x}^{(\ell)}$ denotes the denoising-network layer index, while
the subscript $k$ in $\mathbf{x}_k$ denotes the diffusion step.

\paragraph{Environment feature pyramid.}
The environment input is a three-channel map containing obstacle cells, free cells, and goal cells.
A lightweight convolutional pyramid produces multi-scale feature maps
\begin{equation}
\{\mathbf{F}^{(s)}\}_{s=1}^{S}
=
E_{\mathrm{map}}(\mathbf{G};\mathbf{c}_{\mathrm{g}}),
\qquad
\mathbf{F}^{(s)}\in\mathbb{R}^{D\times H_s\times W_s}.
\end{equation}
Each convolutional stage is modulated by the global condition through FiLM~\citep{perez2017filmvisualreasoninggeneral}:
\begin{equation}
\mathrm{FiLM}(\mathbf{F};\mathbf{c}_{\mathrm{g}})
=
\left(1+\boldsymbol{\gamma}(\mathbf{c}_{\mathrm{g}})\right)\odot \mathbf{F}
+
\boldsymbol{\beta}(\mathbf{c}_{\mathrm{g}}).
\end{equation}
The pyramid uses two strided downsampling stages and a partial upsampling path, and each scale is
projected to the hidden dimension $D$.

\paragraph{Inferred trajectory and sparse social graph.}
At every diffusion step, the network constructs a soft inferred trajectory from
the current noisy action distribution. This trajectory is used only as an
internal continuous estimate for interaction modeling and map-feature sampling,
rather than as a discrete MAPF path. Let $\Delta(a)\in\mathbb{R}^{2}$ be the
grid displacement of action $a$ in row-column coordinates. The expected
displacement is
\begin{equation}
\bar{\Delta}_{i,\tau}
=
\sum_{a=1}^{C}
\mathbf{x}_{k,i,\tau,a}\Delta(a).
\end{equation}
Since start locations are represented in normalized coordinates, the expected
grid displacement is rescaled by the map size before accumulation. The soft
inferred position is then
\begin{equation}
\mathbf{p}^{\mathrm{inf}}_{i,\tau}
=
\mathbf{s}_i
+
\sum_{u=1}^{\tau}
\mathrm{Norm}_{\mathbf{m}}\!\left(\bar{\Delta}_{i,u}\right),
\label{eq:inferred-trajectory}
\end{equation}
where $\mathrm{Norm}_{\mathbf{m}}(\cdot)$ converts row-column grid displacement
into the normalized coordinate system according to the map size
$\mathbf{m}=(W,H)$.

The uncertainty of each action distribution is measured by the normalized entropy
and is later used to enlarge the environment-sensing radius in
Eq.~\eqref{eq:env-sampling-location}:
\begin{equation}
e_{i,\tau}
=
-\frac{1}{\log C}
\sum_{a=1}^{C}
\mathbf{x}_{k,i,\tau,a}
\log \mathbf{x}_{k,i,\tau,a}.
\label{eq:action-entropy}
\end{equation}

The sparse social graph is constructed from these soft inferred trajectories.
For each pair of agents, we compute their minimum inferred-trajectory distance:
\begin{equation}
d_{ij}
=
\min_{\tau=1,\ldots,T}
\left\|
\mathbf{p}^{\mathrm{inf}}_{i,\tau}
-
\mathbf{p}^{\mathrm{inf}}_{j,\tau}
\right\|_1.
\end{equation}
Let $M_k$ be a clipped top-$k$ neighbor count that increases with the noise
level, with the non-self neighbor ratio scheduled from $0.10$ to $0.25$. For
each agent $i$, $\mathcal{N}_k(i)$ contains agent $i$ itself and the $M_k$
nearest non-self agents according to $d_{ij}$. Invalid or padded neighbors are
masked before the attention softmax.

\paragraph{Interaction backbone.}
The backbone consists of $L=4$ interaction blocks. Each block contains temporal attention, sparse
social attention, deformable environment sensing, and a feed-forward network. Adaptive layer
normalization uses the agent-wise condition:
\begin{equation}
\mathrm{AdaLN}(\mathbf{x}_{i,\tau};\mathbf{c}_i)
=
\left(1+\boldsymbol{\gamma}_i\right)\odot
\mathrm{LN}(\mathbf{x}_{i,\tau})
+
\boldsymbol{\beta}_i,
\qquad
[\boldsymbol{\gamma}_i,\boldsymbol{\beta}_i]
=
\psi_{\mathrm{ada}}(\mathbf{c}_i).
\end{equation}
For block $\ell$, the residual updates are
\begin{equation}
\begin{aligned}
\mathbf{x}^{(\ell,1)}
&=
\mathbf{x}^{(\ell)}
+
\mathrm{TempAttn}
\left(
\mathrm{AdaLN}(\mathbf{x}^{(\ell)},\mathbf{c})
\right),
\\
\mathbf{x}^{(\ell,2)}
&=
\mathbf{x}^{(\ell,1)}
+
\mathrm{SocialAttn}
\left(
\mathrm{AdaLN}(\mathbf{x}^{(\ell,1)},\mathbf{c}),
\mathbf{p}^{\mathrm{inf}},
\mathcal{N}_k
\right),
\\
\mathbf{x}^{(\ell,3)}
&=
\mathbf{x}^{(\ell,2)}
+
\mathrm{EnvSense}
\left(
\mathrm{AdaLN}(\mathbf{x}^{(\ell,2)},\mathbf{c}),
\mathbf{p}^{\mathrm{inf}},
\mathbf{e},
\{\mathbf{F}^{(s)}\}_{s=1}^{S},
\boldsymbol{\eta}
\right),
\\
\mathbf{x}^{(\ell+1)}
&=
\mathbf{x}^{(\ell,3)}
+
\mathrm{FFN}
\left(
\mathrm{AdaLN}(\mathbf{x}^{(\ell,3)},\mathbf{c})
\right).
\end{aligned}
\label{eq:interaction-block}
\end{equation}
Here $\mathbf{p}^{\mathrm{inf}}$ and $\mathbf{e}$ denote the soft inferred
positions and normalized action entropies defined above.

The temporal attention module is applied independently to each agent trajectory.
For agent $i$, it treats $\{\mathbf{x}_{i,\tau}\}_{\tau=1}^{T}$ as a temporal
token sequence and performs self-attention along the time dimension. Local
window attention restricts each timestep $\tau$ to attend only to tokens
$u\in\mathcal{W}(\tau)$ within the same window of size $W_t=32$:
\begin{equation}
\mathrm{TempAttn}_{\mathrm{local}}(\mathbf{x}_{i,\tau})
=
\sum_{u\in\mathcal{W}(\tau)}
\mathrm{softmax}_{u}
\left(
\frac{\mathbf{q}_{i,\tau}^{\top}\mathbf{k}_{i,u}}{\sqrt{d}}
\right)
\mathbf{v}_{i,u}.
\end{equation}
The module additionally uses sparse global attention over temporal anchors
sampled every $S_t=16$ timesteps, allowing each token to receive coarse
long-range context without full attention over all $T$ timesteps.

The sparse social attention module operates at each timestep over the dynamic
neighborhood $\mathcal{N}_k(i)$. Let $\mathbf{q}_{i,\tau}$,
$\mathbf{k}_{j,\tau}$, and $\mathbf{v}_{j,\tau}$ be the query, key, and value
projections. The attention weights are
\begin{equation}
\alpha_{ij,\tau}
=
\mathrm{softmax}_{j\in\mathcal{N}_k(i)}
\left(
\frac{
\mathbf{q}_{i,\tau}^{\top}\mathbf{k}_{j,\tau}
}{
\sqrt{d}
}
+
b_{\theta}
\left(
\mathbf{p}^{\mathrm{inf}}_{i,\tau}
-
\mathbf{p}^{\mathrm{inf}}_{j,\tau}
\right)
\right),
\end{equation}
where $b_{\theta}(\cdot)$ is a learned geometric bias from relative inferred
positions. The social update is
\begin{equation}
\mathrm{SocialAttn}_{i,\tau}
=
\sum_{j\in\mathcal{N}_k(i)}
\alpha_{ij,\tau}\mathbf{v}_{j,\tau}.
\end{equation}

The environment-sensing module samples map features around the inferred
trajectory. Let $s=1,\ldots,S$ index the map-pyramid level and
$p=1,\ldots,P$ index the sampling point. For each agent $i$ and timestep
$\tau$, the module predicts a sampling offset $\boldsymbol{\delta}^{p}_{i,\tau}$
and an attention weight $\omega^{s,p}_{i,\tau}$ over scale-point pairs:
\begin{equation}
\boldsymbol{\delta}^{p}_{i,\tau}
=
\tanh\!\left(W_{\delta}^{p}\mathbf{x}_{i,\tau}\right),
\qquad
\omega^{s,p}_{i,\tau}
=
\mathrm{softmax}_{s,p}
\left(
W_{\omega}^{s,p}\mathbf{x}_{i,\tau}
\right).
\end{equation}
The sampling location $\mathbf{u}^{p}_{i,\tau}$ is centered at the inferred
position $\mathbf{p}^{\mathrm{inf}}_{i,\tau}$ and expanded according to the
action-distribution entropy:
\begin{equation}
\mathbf{u}^{p}_{i,\tau}
=
\mathbf{p}^{\mathrm{inf}}_{i,\tau}
+
r_0(1+0.2e_{i,\tau})\boldsymbol{\delta}^{p}_{i,\tau},
\label{eq:env-sampling-location}
\end{equation}
where $r_0$ is a learned base radius and $e_{i,\tau}$ is the normalized entropy
defined in Eq.~\eqref{eq:action-entropy}. With bilinear interpolation
$\mathcal{B}(\mathbf{F}^{(s)},\mathbf{u})$ on the $s$-th map feature
$\mathbf{F}^{(s)}$, the environment context is
\begin{equation}
\mathbf{z}_{i,\tau}
=
\sum_{s=1}^{S}\sum_{p=1}^{P}
\omega^{s,p}_{i,\tau}\,
\eta_s\,
\mathcal{B}\!\left(\mathbf{F}^{(s)},\mathbf{u}^{p}_{i,\tau}\right),
\label{eq:env-sensing}
\end{equation}
where $\eta_s$ is the scale-gate weight from Eq.~\eqref{eq:scale-gate}. The
resulting context $\mathbf{z}_{i,\tau}$ is projected back to dimension $D$ to
form the residual environment-sensing update used in
Eq.~\eqref{eq:interaction-block}.

\paragraph{Output head.}
After the final interaction block, the network normalizes the denoising tokens
and maps each token to action logits through a linear output head:
\begin{equation}
\boldsymbol{\ell}_{i,\tau,:}
=
\mathrm{Linear}_{\mathrm{out}}
\left(
\mathrm{LN}\!\left(\mathbf{x}^{(L)}_{i,\tau}\right)
\right)
\in\mathbb{R}^{C}.
\end{equation}
Collecting all agents and timesteps gives
$f_{\theta}(\mathbf{x}_k,\mathbf{c},k)\in\mathbb{R}^{N\times T\times C}$.
The clean-action probability vector is obtained by normalizing the logits over
the action dimension:
\begin{equation}
\hat{\mathbf{x}}_{0,i,\tau,:}
=
\mathrm{softmax}\!\left(\boldsymbol{\ell}_{i,\tau,:}\right).
\end{equation}
These probability vectors parameterize the factorized clean-state distribution
predicted by the denoiser:
\begin{equation}
\tilde{p}_{\theta}(\mathbf{x}_0\mid \mathbf{x}_k,\mathbf{c})
=
\prod_{i=1}^{N}\prod_{\tau=1}^{T}
\mathrm{Cat}\!\left(
\mathbf{x}_{0,i,\tau,:};
\hat{\mathbf{x}}_{0,i,\tau,:}
\right),
\end{equation}
which is then used to construct the D3PM reverse transition
$p_{\theta}(\mathbf{x}_{k-1}\mid \mathbf{x}_k,\mathbf{c})$.

\subsection{Training Objective}
\label{app:Training Objective}

For completeness, the generic D3PM variational term referenced in
Eq.~\eqref{eq:d3pm-objective} decomposes as
\begin{equation}
\begin{aligned}
\mathcal{L}_{\mathrm{vb}}
=
\mathbb{E}_{q(\mathbf{x}_0)}
\Big[
&\mathrm{KL}\!\left(q(\mathbf{x}_K \mid \mathbf{x}_0)\,\|\,p(\mathbf{x}_K)\right)
+
\sum_{k=2}^{K}
\mathbb{E}_{q(\mathbf{x}_k \mid \mathbf{x}_0)}
\left[
\mathrm{KL}\!\left(
q(\mathbf{x}_{k-1}\mid \mathbf{x}_k,\mathbf{x}_0)
\,\|\,
p_\theta(\mathbf{x}_{k-1}\mid \mathbf{x}_k)
\right)
\right]
\\
&-
\mathbb{E}_{q(\mathbf{x}_1 \mid \mathbf{x}_0)}
\left[
\log p_\theta(\mathbf{x}_0 \mid \mathbf{x}_1)
\right]
\Big].
\end{aligned}
\label{eq:app-d3pm-vb}
\end{equation}
In the conditional MAPF setting, let $q(\mathbf{x}_0,\mathbf{c})$ denote the
expert data distribution over clean joint action tensors and MAPF conditions.
Our main text uses the generative objective
\begin{equation}
\mathcal{L}_{\mathrm{gen}}
=
\mathcal{L}_{\mathrm{aux}}
+
\lambda_{\mathrm{KL}}\mathcal{L}_{\mathrm{KL}},
\label{eq:app-lgen}
\end{equation}
where $\mathcal{L}_{\mathrm{aux}}$ is the auxiliary clean-state prediction term
in Eq.~\eqref{eq:d3pm-objective}, and $\mathcal{L}_{\mathrm{KL}}$ is the
posterior-matching component extracted from the variational term
$\mathcal{L}_{\mathrm{vb}}$. Instead of evaluating the full variational
decomposition, we keep the intermediate KL term that aligns the learned reverse
transition with the analytic discrete posterior:
\begin{equation}
\mathcal{L}_{\mathrm{KL}}
=
\mathbb{E}_{q(\mathbf{x}_0,\mathbf{c})}
\mathbb{E}_{k\sim\mathrm{Unif}(\{1,\ldots,K\})}
\mathbb{E}_{q(\mathbf{x}_k\mid \mathbf{x}_0)}
\left[
\mathrm{KL}\!\left(
q(\mathbf{x}_{k-1}\mid \mathbf{x}_k,\mathbf{x}_0)
\,\|\, 
p_{\theta}(\mathbf{x}_{k-1}\mid \mathbf{x}_k,\mathbf{c})
\right)
\right].
\label{eq:app-kl-loss}
\end{equation}
Under the $\mathbf{x}_0$-parameterization,
$p_{\theta}(\mathbf{x}_{k-1}\mid \mathbf{x}_k,\mathbf{c})$ is induced by the
predicted clean-state distribution
$\tilde{p}_{\theta}(\mathbf{x}_0\mid \mathbf{x}_k,\mathbf{c})$ through the
analytic D3PM posterior, as in Eq.~\eqref{eq:d3pm-reverse}.

We further add the task-oriented auxiliary objective used in the main text:
\begin{equation}
\mathcal{L}
=
\mathcal{L}_{\mathrm{gen}}
+
\mathcal{L}_{\mathrm{task}},
\label{eq:app-total-loss}
\end{equation}
with
\begin{equation}
\mathcal{L}_{\mathrm{task}}
=
\lambda_{\mathrm{goal}}\mathcal{L}_{\mathrm{goal}}
+
\lambda_{\mathrm{vertex}}\mathcal{L}_{\mathrm{vertex}}
+
\lambda_{\mathrm{edge}}\mathcal{L}_{\mathrm{edge}}
+
\lambda_{\mathrm{valid}}\mathcal{L}_{\mathrm{valid}}.
\label{eq:app-task-loss}
\end{equation}
The default weights are
\[
\lambda_{\mathrm{KL}}=0.02,\quad
\lambda_{\mathrm{goal}}=0.4,\quad
\lambda_{\mathrm{vertex}}=0.2,\quad
\lambda_{\mathrm{edge}}=0.2,\quad
\lambda_{\mathrm{valid}}=0.4.
\]
The task losses are computed from the predicted clean-action distribution
$\hat{\mathbf{x}}_0$ through its induced soft trajectory, rather than from
samples generated by the full reverse diffusion process. Therefore,
$\mathcal{L}_{\mathrm{task}}$ serves as an auxiliary shaping signal, while
$\mathcal{L}_{\mathrm{gen}}$ remains the primary expert-imitation objective. We
next define each task-oriented term used in Eq.~\eqref{eq:app-task-loss}.

\paragraph{Goal-progress loss.}
We use a BFS-based goal-progress loss. For each agent and timestep, let
$d_{i,\tau}$ be the shortest-path distance from the predicted position to the
goal on the obstacle map, and let $\tilde d_{i,\tau}=d_{i,\tau}/s$ be the
normalized distance, where $s=\max(H_{\mathrm{map}},W_{\mathrm{map}})$ is the
map scale. For trajectory horizon $T$, the loss penalizes failures to make
progress toward the goal:
\begin{equation}
\begin{aligned}
\mathcal{L}_{\mathrm{goal}}
=
\frac{1}{N(T-1)}
\sum_{i=1}^{N}\sum_{\tau=0}^{T-2}
\max\Bigl(
&\tilde d_{i,\tau+1}
-
\max\bigl(\tilde d_{i,\tau}-\tfrac{1}{s},\,0\bigr),
\,0
\Bigr).
\end{aligned}
\end{equation}
This objective encourages each step to reduce the shortest-path distance by
approximately one cell whenever possible.

\paragraph{Vertex-conflict loss.}
Let $\mathbf{p}_{i,\tau}$ denote the expected position of agent $i$ at timestep
$\tau$, obtained from the predicted clean-action probabilities
$\hat{\mathbf{x}}_0$. We penalize agents that come within a Manhattan safety
radius $r_{\mathrm{v}}$:
\begin{equation}
\mathcal{L}_{\mathrm{vertex}}
=
\frac{1}{|\mathcal{S}_{\mathrm{v}}|}
\sum_{(i,j,\tau)\in\mathcal{S}_{\mathrm{v}}}
\exp\!\left(
-\|\mathbf{p}_{i,\tau}-\mathbf{p}_{j,\tau}\|_1
\right),
\end{equation}
where
\begin{equation}
\mathcal{S}_{\mathrm{v}}
=
\left\{
(i,j,\tau): i\neq j,\,
\|\mathbf{p}_{i,\tau}-\mathbf{p}_{j,\tau}\|_1 \le r_{\mathrm{v}}
\right\}.
\end{equation}

\paragraph{Edge-conflict loss.}
We penalize swap-like conflicts, where two agents move through the same edge in
opposite directions between consecutive timesteps. For agents $i$ and $j$, this
occurs when $\mathbf{p}_{i,\tau}$ is close to $\mathbf{p}_{j,\tau+1}$ and
$\mathbf{p}_{i,\tau+1}$ is close to $\mathbf{p}_{j,\tau}$. We therefore define
\begin{equation}
d^{\rightarrow}_{ij,\tau}
=
\|\mathbf{p}_{i,\tau}-\mathbf{p}_{j,\tau+1}\|_1,
\qquad
d^{\leftarrow}_{ij,\tau}
=
\|\mathbf{p}_{i,\tau+1}-\mathbf{p}_{j,\tau}\|_1 .
\end{equation}
Let $\mathcal{S}_{\mathrm{e}}$ contain all triples $(i,j,\tau)$ for which both
distances are no larger than the edge-conflict radius $r_{\mathrm{e}}$. The
edge-conflict loss is
\begin{equation}
\mathcal{L}_{\mathrm{edge}}
=
\frac{1}{|\mathcal{S}_{\mathrm{e}}|}
\sum_{(i,j,\tau)\in\mathcal{S}_{\mathrm{e}}}
\exp\!\left(
-\frac{
d^{\rightarrow}_{ij,\tau}
+
d^{\leftarrow}_{ij,\tau}
}{2}
\right).
\end{equation}

\paragraph{Action-validity loss.}
Let $\mathcal{A}_{i,\tau}^{\mathrm{valid}}$ be the set of actions that remain
in bounds and do not enter obstacles from the current rollout position. We
penalize probability mass assigned to invalid actions:
\begin{equation}
\mathcal{L}_{\mathrm{valid}}
=
\frac{1}{NT}
\sum_{i,\tau}
\sum_{a\notin \mathcal{A}_{i,\tau}^{\mathrm{valid}}}
\hat{x}_{0,i,\tau,a}.
\end{equation}

\subsection{Training Schedule}

Training follows a two-stage schedule. In the first stage, we optimize only
the generative objective $\mathcal{L}_{\mathrm{gen}}$, which consists of the
clean-state auxiliary loss and the diffusion posterior-matching KL loss. The
task-oriented losses are disabled during the first 60 epochs. In the second
stage, we gradually introduce the task-oriented objective
$\mathcal{L}_{\mathrm{task}}$ over a 150-epoch warmup period. Let $e$ denote the
number of epochs elapsed after the first stage. The task-loss scaling factor is
\begin{equation}
\alpha(e)
=
0.2
+
0.8\cdot
\min\!\left(\frac{e}{150},\,1\right).
\end{equation}
The weights of $\mathcal{L}_{\mathrm{goal}}$,
$\mathcal{L}_{\mathrm{vertex}}$, $\mathcal{L}_{\mathrm{edge}}$, and
$\mathcal{L}_{\mathrm{valid}}$ are multiplied by $\alpha(e)$ during this
warmup.

\subsection{Optimization Hyperparameters}

We optimize the model with AdamW~\citep{loshchilov2019decoupledweightdecayregularization} using a constant learning rate of
$2\times 10^{-4}$ and weight decay $10^{-4}$. Gradients are clipped to norm
$1.0$. We do not use a learning-rate scheduler or an exponential moving average.
The denoiser uses hidden dimension 128, conditioning dimension 128, 4 attention
heads, 4 interaction layers, and dropout 0. The deformable environment-sensing
module uses 8 sampling points per head and 3 feature scales.

\begin{table}[t]
\centering
\small
\setlength{\tabcolsep}{8pt}
\renewcommand{\arraystretch}{1.12}
\caption{Main supervised training hyperparameters used by DiffLNS.}
\label{tab:training-hparams}
\begin{tabular}{lc}
\toprule
\textbf{Hyperparameter} & \textbf{Value} \\
\midrule
Training map size & $23\times23$ \\
Training map categories & maze, room, warehouse \\
Training agent counts & $N\in\{32,64,96\}$ \\
Dataset mixing mode & round-robin across fixed-$N$ subsets \\
Per-GPU batch size & 8 \\
Number of training epochs & 400 \\
Generative-only stage & 60 epochs \\
Task-loss warmup & 150 epochs \\
Optimizer & AdamW \\
Learning rate & $2\times10^{-4}$ \\
Weight decay & $10^{-4}$ \\
Gradient clipping & 1.0 \\
Precision & BF16 autocast \\
Random seed & 42 \\
Diffusion steps & 100 \\
Beta schedule & cosine \\
Number of actions & 5 \\
$\lambda_{\mathrm{KL}}$ & 0.02 \\
$\lambda_{\mathrm{goal}}$ & 0.4 \\
$\lambda_{\mathrm{vertex}}$ & 0.2 \\
$\lambda_{\mathrm{edge}}$ & 0.2 \\
$\lambda_{\mathrm{valid}}$ & 0.4 \\
Hidden dimension & 128 \\
Condition dimension & 128 \\
Attention heads & 4 \\
Interaction layers & 4 \\
Dropout & 0 \\
Temporal window $W_t$ & 32 \\
Temporal global stride $S_t$ & 16 \\
Environment sampling points & 8 \\
Environment feature scales & 3 \\
\bottomrule
\end{tabular}
\end{table}

\end{document}